\documentclass[10pt,twocolumn,letterpaper]{article}
\usepackage[final]{cvpr}      

\definecolor{cvprblue}{rgb}{0.21,0.49,0.74}
\usepackage[pagebackref,breaklinks,colorlinks,allcolors=cvprblue]{hyperref}

\def\paperID{21623} 
\def\confName{CVPR}
\def\confYear{2026}
\usepackage[T1]{fontenc}
\usepackage{graphicx}
\usepackage{booktabs}
\usepackage[misc]{ifsym}
\usepackage{amsmath}
\usepackage{pifont}
\usepackage{xcolor}
\usepackage[table,xcdraw]{xcolor}
\usepackage{textgreek}

\usepackage{subcaption}
\usepackage{float}
\usepackage{graphicx}
\usepackage{subcaption}
\usepackage{wrapfig}
\usepackage{multirow}
\usepackage{arydshln}
\usepackage{enumitem}
\usepackage{subcaption}

\usepackage{mwe}

\newcommand{\cmark}{\ding{51}}%

\newcommand{\system}{{\textsl{AdaBet}}} 

\newtheorem{definition}{Definition}

\title{\system: Gradient-free Layer Selection  for Efficient Training \\ of Deep Neural Networks}

\begin{document}


\author{}

\author{
Irene Tenison$^{1}$\thanks{Work mainly done during an internship at Nokia Bell Labs. Emails: itenison@mit.edu; sjituit@gmail.com; 
fahim.kawsar@glasgow.ac.uk
},
Soumyajit Chatterjee$^{2}$,
Fahim Kawsar$^{2,3}$,
Mohammad Malekzadeh$^{2}$ \\
$^{1}$MIT, Cambridge, MA, USA \quad
$^{2}$Nokia Bell Labs, Cambridge, UK \quad
$^{3}$University of Glasgow, UK\\
}

\maketitle 
\begin{abstract}
To utilize pre-trained neural networks on edge and mobile devices, we often require efficient adaptation to user-specific runtime data distributions while operating under limited compute and memory resources. On-device retraining with a target dataset can facilitate such adaptations; however, it remains impractical due to the increasing depth of modern neural nets, as well as the computational overhead associated with gradient-based optimization across all layers. Current approaches reduce training cost by selecting a subset of layers for retraining; however, they rely on labeled data, at least one full-model backpropagation, or server-side meta-training, limiting their suitability for constrained devices. We introduce \system{}, a gradient-free layer selection approach to rank important layers, followed by important channels of these layers, by analyzing topological features of their activation spaces through Betti Numbers and using forward passes alone. \system{} allows selecting layers and channels with high learning capacity, which are important for retraining and adaptation, without requiring labels or gradients. Evaluating \system{} on sixteen pairs of benchmark models and datasets shows \system{} achieves an average gain of 2.5\% more classification accuracy over gradient-based baselines while reducing average peak memory consumption by 40\%. We open-source our code at 
\url{https://github.com/Nokia-Bell-Labs/efficient_layer_selection}
\end{abstract}


\section{Introduction}

Consider waking up to our mobile or wearable device that has fine‑tuned, overnight, and entirely on‑device, a machine learning~(ML) model to our unique skin tone and texture, personalizing melanoma screening without transmitting a single image to the cloud~\cite{bdair2021fedperl, swanson2023patterns}. By locally retraining a pre-trained ML model, we create specialized apps that preserve user privacy while providing personalized services for individual needs. Such an approach can extend from real-time mood prediction based on subtle facial microexpressions to adaptive glucose-monitoring algorithms on insulin pumps, each model continuously refining itself at the user's side. Since on-device training still faces key challenges in resource consumption and computational efficiency, maintaining the right balance between performance and resource use is essential to enabling the next generation of efficient, privacy-preserving, and user-centric ML technologies~\cite{laskaridis2024future}.

In particular, deep neural networks (DNNs) deployed on edge or personal devices are typically pre-trained on large, generic datasets. For many apps, it is often necessary to retrain or fine-tune the DNN using a new, small dataset to enhance performance on a desired task. Such retraining, when performed locally on the device, enables personalization and preserves user privacy by avoiding interactions with a server. A plethora of applications benefit from such local training: perception systems in autonomous vehicles~\cite{sun2020scalability}, wildlife monitoring via camera traps~\cite{10.1073/pnas.1719367115}, personalized speech recognition~\cite{10.5555/3618408.3619590}, and health monitoring via wearables~\cite{battery_3}.

The problem remains the intensive computational requirements, substantial memory usage, and high power consumption associated with gradient-based training, making it impractical for many resource-constrained scenarios. Unlike cloud servers with dedicated GPUs and abundant memory, edge devices operate under strict limitations on memory, compute throughput, and battery life~\cite{battery_1,battery_2,battery_3,ElasticTrainer}. To train DNNs, we require both forward and backward passes through all layers, which significantly increases memory usage and compute time compared to inference-only tasks. The backward pass typically consumes 3$\times$ more memory than the forward pass due to gradient and activation storage~\cite{tinytl}, often making full backpropagation infeasible on some devices. Therefore, enabling practical on-device learning demands novel approaches that prioritize memory efficiency and computational reduction while maintaining the model's accuracy.

An approach to these challenges is to adopt a shallow neural network architecture. This offers a straightforward and memory-efficient solution, but suffers from limited capacity, which can hinder performance on some tasks. Another approach is retraining a DNN using  techniques such as gradient checkpointing~\cite{gradient_checkpointing,chen2016training}, layer-wise training~\cite{bengio2006greedy}, split learning~\cite{9835178}, or Transfer Learning-based methods~\cite{jiang2022back}. While memory-efficient, these solutions introduce trade-offs: gradient checkpointing reduces memory usage at the cost of compute overhead and increased running time~\cite{singh2024study}, and other methods like layer-wise training, split-learning, and Transfer Learning often suffer from suboptimal performance when there are significant changes in data distribution or downstream tasks~\cite{lee2023surgical,sakamoto2024end}. Moreover, some require communication overhead if parts of the model need to be offloaded to external compute platforms~\cite{9835178}. A third approach focuses on hardware-specific design strategies, such as quantization-aware training~\cite{ashkboos2024efqat} and hardware-aware neural architecture search~\cite{10.1145/3437984.3458836}, which are primarily optimized for inference rather than for flexible or on-device training.

A recent approach involves using sparse updates, where only the most important layers are updated to save memory, an idea presented by the Tiny Training Engine~\cite{256kb}. However, it performs layer selection at the server side and does not support dynamic, on-device adaptation when the data distribution shifts. Selective fine-tuning strategies preserve the generalized and well-aligned layers of the pre-trained model and adapt only the mismatched layers. This acts as an implicit regularization and often leads to higher accuracies. TinyTrain~\cite{TinyTrain} and ElasticTrainer~\cite{ElasticTrainer} extend sparse updates to support on-device, task-aware re-training. TinyTrain achieves this by computing Fisher Information to select important parameters, but requires at least one full round of backpropagation through the entire model and relies on server-side meta-training. ElasticTrainer uses a dynamic programming approach to select parameters. While offering more adaptive training, ElasticTrainer requires labeled data and multiple selection steps.

\begin{figure}[t]
\centering
\includegraphics[width=.4\textwidth,keepaspectratio]{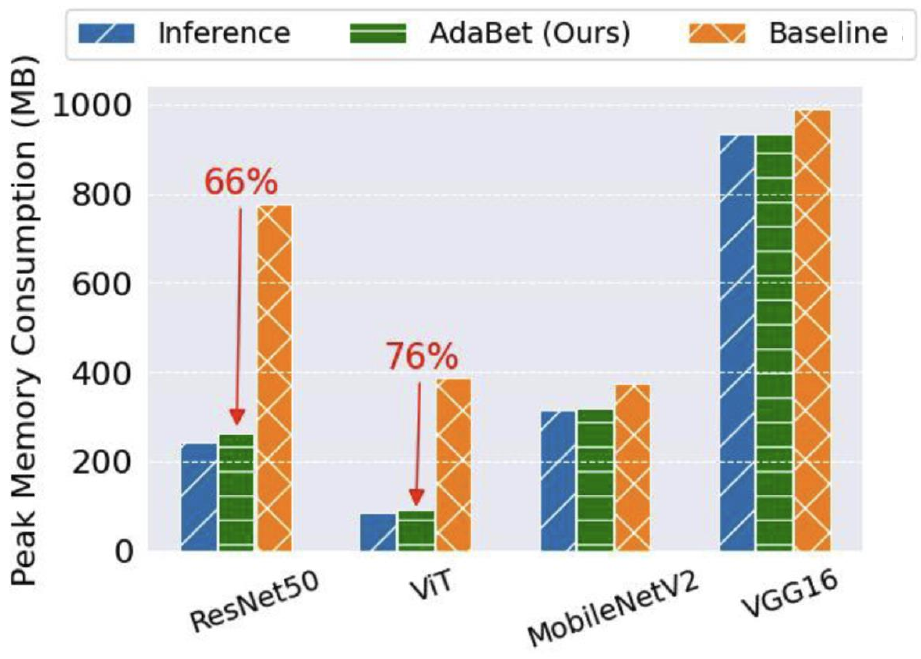}
\caption{\emph{Memory efficiency.} \system{} reduces peak memory consumption of retraining by up to 76\% (on average by 40\% for different pre-trained DNNs retrained on Oxford-IIIT Pets), with peak memory of inference-only and full-model training as baselines.
}
    \label{fig:mem-1}
\end{figure}

\sloppy\textbf{Our Contribution.} We propose \system{}, a memory- and compute-efficient framework for retraining of DNNs. The main novelty of \system{} is \emph{gradient-free} layer selection, eliminating the need for either full-model backpropagation or server-side meta-training. To achieve this, \system{} selects trainable layers based on their learning capacity, quantified via topological features derived from the first Betti Number of each layer’s activation outputs~\cite{topological_features}. Betti Numbers are achieved during a forward pass and are normalized by the size of each layer to align with memory efficiency objectives. As this process is independent of label availability, \system{} is suitable for even unsupervised scenarios. We benchmark \system{} across several benchmarks using pre-trained DNNs: ResNet, VGG16, MobileNet, and ViT. Results show that \system{} achieves comparable or better accuracy than gradient-based methods while reducing average peak memory usage by 40\% (shown in \figurename~\ref{fig:mem-1}) and obtaining an average accuracy gain of +5\% (detailed in \S\ref{sec_results}).
In summary, the key contributions are:
\begin{enumerate}

    \item A memory- and compute-efficient on-device (or {under-budget}) training enabled by {gradient-free} layer and channel selection without requiring labeled data or server-side meta training.
    
    \item Leveraging topological features based on well-studied Betti Numbers of activations, for selecting a subset of layers and channels based on their learning capacity and resource requirements.
    
    \item Providing better balance between resource efficiency and accuracy, benchmarked on popular pre-trained DNNs across a range of image classification tasks.
    
\end{enumerate}

\section{Related Work: On-device Layer Selection}
\label{rel_work}
The demand for on-device training to support personalization and privacy necessitates local backpropagation, which is resource-intensive due to the computation and storage of gradients. Existing strategies, such as gradient checkpointing~\cite{gradient_checkpointing,chen2016training}, layer-wise training~\cite{bengio2006greedy}, split learning~\cite{9835178}, and transfer learning~\cite{jiang2022back}, reduce computation but often degrade accuracy under distribution or task shifts. Other approaches jointly optimize accuracy and resource usage~\cite{256kb,tinytl,E2Train,Adadrop,10.1145/3437984.3458836}. However, they either require server-side architecture search prior to deployment or perform at least one full backpropagation, which remains costly for constrained devices. PruneTrain~\cite{PruneTrain} reduces training cost via structured pruning. TinyTrain~\cite{TinyTrain} selects layers using Fisher Information but depends on a full backward pass and an additional meta-training stage, which may become ineffective under distribution mismatch~\cite{FI_ref1,FI_ref2,FI_ref3}. ElasticTrainer~\cite{ElasticTrainer} eliminates the need for meta-training by performing dynamic programming-based tensor selection directly on-device; however, this process involves multiple full backpropagation rounds, often resulting in memory overhead comparable to that of full training. 
A comparison of these methods and our proposed \system{} is provided in Table~\ref{tbl:related_work} and a detailed analysis of related works is given in Appendix \ref{Supp_A}

\begin{table}[t]
\centering
\caption{Compared to related work, our \system{} is the only gradient-free (i.e., label-free) selection method. \system{} does not require server-side meta-training.
}
\label{tbl:related_work}
\resizebox{\columnwidth}{!}{%
\begin{tabular}{l|cccc|l} 
\rowcolor[rgb]{1.,1.,1.} & \begin{tabular}[c]{@{}>{\cellcolor[rgb]{1.,1.,1.}}c@{}}Unlabeled\\data? \end{tabular} & \begin{tabular}[c]{@{}>{\cellcolor[rgb]{1.,1.,1.}}c@{}}Has Layer \\ selection?\end{tabular} & \begin{tabular}[c]{@{}>{\cellcolor[rgb]{1.,1.,1.}}c@{}} One-step \\ selection\end{tabular} & \begin{tabular}[c]{@{}>{\cellcolor[rgb]{1.,1.,1.}}c@{}}Server\\ independent?\end{tabular} & \begin{tabular}[c]{@{}>{\cellcolor[rgb]{1.,1.,1.}}c@{}}Selection\\approach\end{tabular} \\ 
\hline
Transfer Learning~\cite{jiang2022back} & $\times$ & $\times$ & $\times$ & \cmark & N/A \\
\rowcolor[rgb]{0.937,0.937,0.937} \begin{tabular}[c]{@{}>{\cellcolor[rgb]{0.937,0.937,0.937}}c@{}}Quant-Aware Training~\cite{ashkboos2024efqat}\end{tabular} & $\times$ & $\times$ & $\times$ & $\times$ & N/A \\
TinyTrainingEngine~\cite{256kb} & $\times$ & \cmark & $\times$ & $\times$ & \begin{tabular}[c]{@{}c@{}}Evolutionary\end{tabular}  \\
\rowcolor[rgb]{0.937,0.937,0.937} TinyTrain~\cite{TinyTrain} & $\times$ & \cmark & $\times$ & $\times$ & Fisher Info \\
PruneTrain~\cite{PruneTrain} & $\times$ & \cmark &  $\times$ & \cmark & \begin{tabular}[c]{@{}c@{}}Lasso Penalty\end{tabular} \\
\rowcolor[rgb]{0.937,0.937,0.937} 
ElasticTrainer~\cite{ElasticTrainer} & $\times$ & \cmark &  $\times$ & \cmark & \begin{tabular}[c]{@{}c@{}}Dynamic Prog.\end{tabular} \\\hline
 \textbf{\system{} (Our)}& \cmark & \cmark & \cmark & \cmark & \begin{tabular}[c]{@{}c@{}}Betti Numbers\end{tabular} \\
\end{tabular}
}
\end{table}


\section{Backgrounds and Motivations}
To enable practical and effective layer selection on-device, it is crucial to identify selection metrics that (1) indicate a layer or channel's learning capacity so that the selection does not hamper performance drastically; (2) operate without the need for gradients (or backpropagation) so that the selection can be performed under resource budgets or on-device unlike existing works; and (3) is server-independent for additional privacy. As shown in Table~\ref{tbl:related_work}, while previous works like ElasticTrainer\cite{ElasticTrainer} has used dynamic programming and TinyTrain\cite{TinyTrain} has used statistical criteria such as Fisher Information. These require backpropagation and labeled data,  limiting their practical deployment on edge devices. Recent advances in topological data analysis suggest that topological invariants, like Betti numbers, may offer unique advantages in assessing the complexity and expressivity of learned representations in neural networks. Motivated by these insights, we next introduce Betti numbers as a powerful tool for gradient-free, label-free layer selection.

\subsection{Betti Numbers}
Topology studies properties of spaces that remain invariant under continuous deformations. In machine learning, this provides a way to analyze the robustness and structure of learned representations beyond visual inspection. Homology characterizes a space in terms of its ``holes'' of different dimensions, capturing connected components, loops, and higher-dimensional voids. Persistent homology enables computing these topological features across scales, making it suitable for analyzing complex activation spaces in DNNs. The key summary statistic used in this work is the $n$\textsuperscript{th} Betti number, denoted by $b_n$, which counts the number of independent $n$-dimensional holes in a space (for example, $b_0$ reflects connected components and $b_1$ reflects loops), as detailed in Appendix \ref{Supp_B} alongside more background on algebraic topology, and homology. Betti numbers, therefore, provide a compact and algebraic description of structural complexity and are robust to transformations such as scaling, rotation, and noise, making them a helpful tool for characterizing representation geometry in deep models. Building on such topological insights, \system{} leverages {\em the first Betti Numbers of each layer’s activation outputs} in a DNN to quantify its complexity, thereby guiding the selection of layers for targeted retraining (as detailed in \S\ref{sec:method}). 

\begin{definition}
Given a set of activations $\mathcal{A}_i=\{a_i\}_{d\in D}$ produced by  layer $i$ over a batch of $D$ inputs, $b_1$ (the first Betti number) of layer $i$ can be defined and computed as the number of persistent one-dimensional topological features (loops) in this complex (more details in Appendix \ref{Supp_B}). 
\end{definition}

\begin{figure}[t]
\centering
\includegraphics[width=0.4\textwidth,keepaspectratio]{images_mobicom/dbscan_acc.pdf}
\caption{Layer-wise Betti Numbers computed on activations of a pre-trained VGG-16 model for Oxford-IIIT Pets, along with the number of clusters obtained via DBSCAN on UMAP reduced layer embeddings and accuracy from isolated training of the layer; all averaged across 5 independent runs. DBSCAN of the embeddings shows a similar ranking pattern, but Betti Numbers offer a more granular ranking of the layers; and both exhibit similar patters with isolated update accuracy.}
\label{fig:dbscan}
\end{figure}

\begin{figure*}[t]
    \centering
    \begin{subfigure}[t]{0.33\textwidth}
        \centering
        \includegraphics[width=\linewidth,keepaspectratio]{images_mobicom/betti_vs_fisher_seeds_3.pdf}
    \end{subfigure}\hfill
    \begin{subfigure}[t]{0.33\textwidth}
        \centering
        \includegraphics[width=\linewidth]{images_mobicom/betti_vs_fisher_seeds_1.pdf}
    \end{subfigure}
    \begin{subfigure}[t]{0.33\textwidth}
        \centering
        \includegraphics[width=\linewidth]{images_mobicom/betti_vs_fisher_seeds_2.pdf}
    \end{subfigure}\hfill
    \caption{Fisher Information vs. Betti Numbers. Pre-trained VGG16  adapted to Stanford Dogs dataset. Green/dark rectangles show selected layers. (a) FI ranking depends on the number of backpropagations over the entire model. (b) FI ranking depends on the batch of data selected (due to changes in the random seed), with considerable changes in accuracy. (c) Our gradient-free  Betti Number ranking of \system{}, is more consistent across different batches of data used to select the layers with negligible changes to the overall accuracy.}\label{fig:layer_selection_motivation}
    \vspace{-0.3cm}
\end{figure*}

\subsubsection{Betti Numbers \& Learning Capacity}
\label{sec:methodology_aggregation}

Unlike existing layer selection methods~\cite{ElasticTrainer,TinyTrain} that require at least one round of backpropagation through all layers, our solution only looks into the topological features obtained from layers' activations in the forward pass. Specifically, we choose the \emph{first Betti Number}, $b_1$, of the activations from all layers in any pre-trained model on a batch of local $\mathbb{D}$ as part of our \emph{selection metric}. In our observations, the zeroth Betti Number, $b_0$, has consistently matched the batch size used during the forward pass. 
On the other hand, the first Betti Number has been shown to quantify the learning capacity of DNN layers~\cite{sun2016depth,topological_features}. Specifically, the first Betti Number, $b_1$, measures the number of one-dimensional holes (loops or tunnels) in the output activation space, and is directly linked to the Rademacher Average capacity of a layer~\cite{corneanu2019does}. Since higher capacity is generally associated with an increased risk of overfitting, such layers may be more suitable candidates for targeted fine-tuning. 

Intuitively, when fine-tuning pre-trained models, the  challenge is to preserve generalizable features while selectively adapting task-specific ones. The topology of the embedding space indicates its quality for the downstream task. Layers with low $b_1$ ($\approx$ simple topology) give disentangled representations that are nearly linearly separable; updating these stable layers risks injecting noise that hurts generalization. Conversely, layers with high $b_1$ ($\approx$ complex topology) give looped structures (entangled manifolds), suggesting that the pre-trained features are not aligned with the downstream task. These key layers with higher $b_1$ can be updated for disentanglement and improved separability, and higher accuracy gains.

Empirically, Betti profiles often peak in early and late layers (\figurename~\ref{fig:layer_selection_motivation}(c)); DBSCAN-based estimates of cluster counts track this pattern (\figurename~\ref{fig:dbscan}), suggesting that Betti numbers capture representational capacity in line with prior observations on cluster separability and expressivity~\cite{sun2016depth,topological_features} as well as accuracy under isolated adaptation. Moreover, Betti Numbers are more robust to small perturbations and noise applied to the topological space~\cite{duarte2022betti}. Since the calculation of Betti Numbers can be done using activations from forward propagation of any randomly selected batch from $\mathbb{D}$, it does not require resource-intensive gradient computations for backpropagation, and also does not need labels.

\subsection{Betti Number vs. Fisher Information}
Given a batch of $N$ samples, and based on the gradients $g$ of the loss function with respect to the activations $a$ of the model, Fisher Information~(FI) is defined as $\Delta =\frac{1}{2 N} \sum_{n=1}^{N}\left( a_{n} g_{n }\right)^{2}$.

TinyTrain~\cite{TinyTrain} uses FI to assess the importance of layers. We outline the drawbacks of FI approach to highlight how Betti Numbers can offer a more efficient and yet more effective alternative for layer selection.

\textbf{Advantages of Betti numbers over FI:} FI-based layer selection requires at least one full backpropagation and labeled data, limiting practicality on constrained devices or label-scarce settings. We observe that it is unstable - the selected subset of layers varies with the number of backpropagations computed before the gradients are collected for calculation of FI as well as across batches due to random seeds (\figurename~\ref{fig:layer_selection_motivation}-(a),(b)), reflecting that early gradients are often misaligned with the target distribution and can bias toward (pre- or) meta-training data; moreover, there is no principled convergence criterion, with peak accuracy sometimes appearing mid-training. In contrast, Betti numbers are computed from forward activations without labels or backpropagation and yield consistent layer rankings across batches, leading to stable accuracy with negligible variance 
(\figurename~\ref{fig:layer_selection_motivation}-(c)). 

\section{
Gradient-free Layer Selection}\label{sec:method}

\subsection{Layer Selection}
Let us consider on-device retraining of a deep neural net, e.g., for model adaptation or personalization. Let $\mathcal{M}$ be a DNN {\em pre-trained} on a global dataset using any type of objective function. For example, a pre-trained model published on HuggingFace, comprising multiple {\em layers}, each containing a varying number of parameters. We assume $\mathcal{M}$ is deployed on a constrained device hosting a local training dataset $\mathbb{D}$, without imposing any restrictions on the characteristics of $\mathbb{D}$. Specifically,  $\mathbb{D}$ can be unlabeled or from a different domain than the pre-training dataset. Our objective is to facilitate on-device re-training of $\mathcal{M}$ with $\mathbb{D}$, taking into account resource efficiency and target task accuracy. To achieve our objective, we should efficiently {\em select} the most important layers from $\mathcal{M}$, and only {\em selected layers} are later retrained on $\mathbb{D}$ using the target task objective.

\subsection{\system{} Overview}
We introduce a novel strategy for selecting {\em most important layers} to retrain, while satisfying three key properties: (a) \system{} is gradient-free, reducing on-device memory and computational requirements; (b) \system{} is label-free, enhancing its generalizability across diverse ML tasks; and (c) \system{} is server-free, making it well-suited for privacy-preserving and fully on-device learning scenarios. 

\begin{figure}
    \centering
    \includegraphics[width=\columnwidth,keepaspectratio]{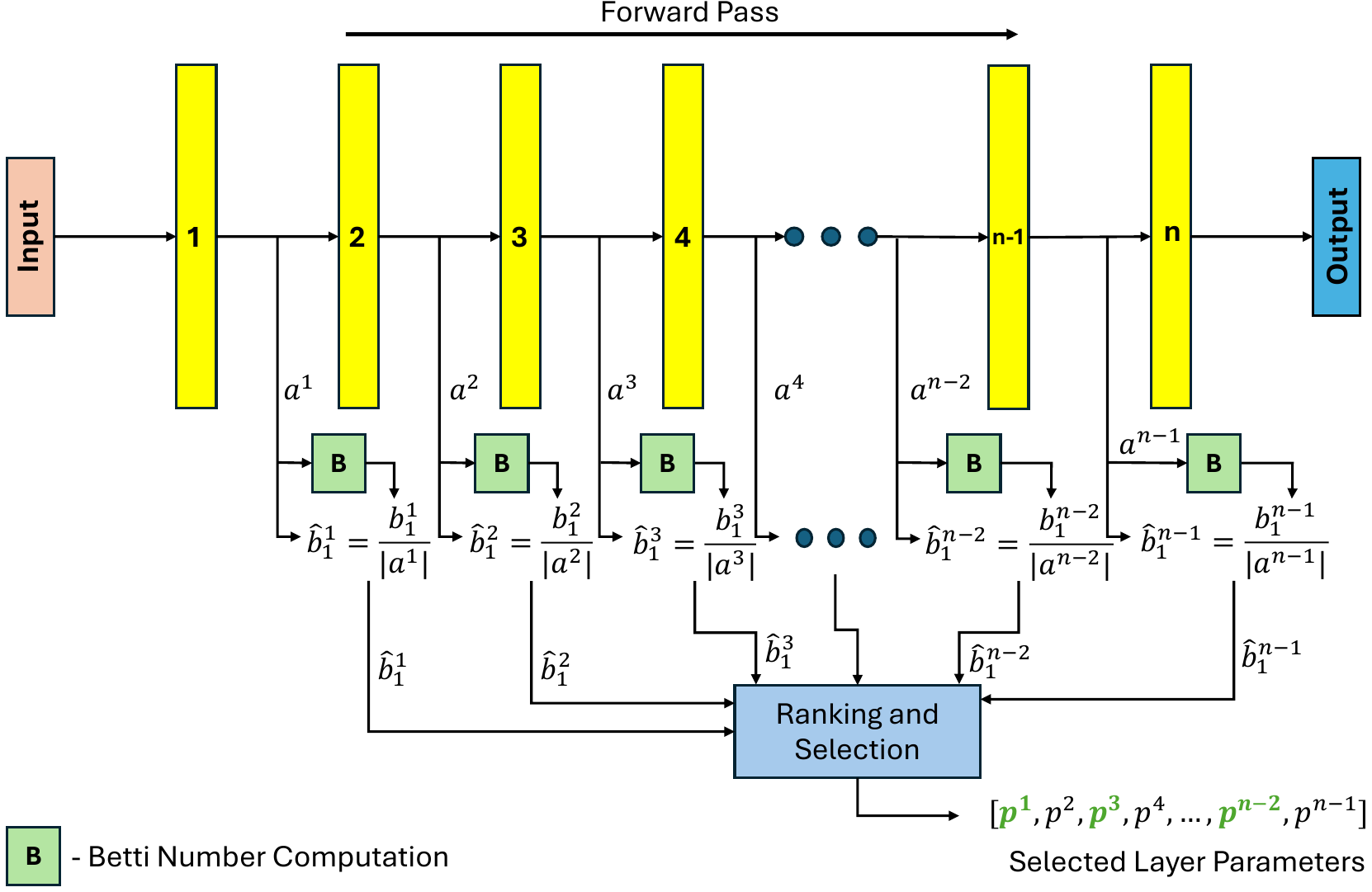}
    \caption{Detailed pipeline of \system{} during its selection phase. Here $a^i$ denotes the activations, $b_{1}^i$ denotes the Betti Number $b_1$ from the activations, $\hat{b}_1^i$ is the normalized Betti Number and $p^i$ is the parameters (weights and biases) of the $i$\textsuperscript{th} layer, respectively.}
    \label{fig:adabet_detailed}
\end{figure}

As illustrated in \figurename~\ref{fig:adabet_detailed}, \system{} comprises three steps: (1) the estimation of each layer's learning capacity via Betti Numbers computed on a subset of $\mathbb{D}$, (2) the selection of important layers based on learning capacity and the size of activations, considering the available computation budget on the device, and (3) the retraining of selected layers in $\mathcal{M}$ using the local dataset $\mathbb{D}$. 

In particular, since $\mathcal{M}$ is pre-trained, we first evaluate all layers' learning capacity by forward propagating samples drawn from $\mathbb{D}$ through the DNN, and we capture the corresponding activations of each layer (see \figurename~\ref{fig:adabet_detailed}). A topological analysis of these activations, via Betti Number calculation, is performed to assess the learning capacity of each layer. The important layers are then selected based on their ranking: layers with higher Betti Numbers are favored, while those with a larger number of parameters are penalized in the selection process. Finally, the re-training of $\mathcal{M}$ by only updating the selected layers using $\mathbb{D}$.

\subsubsection{Batch size flexibility.}\label{sec:aggregation} Small batch sizes (e.g., 4 samples at a time) can make Betti Numbers unreliable, because each small batch gives only a few activation samples. Gradient-based methods, such as ElasticTrainer~\cite{ElasticTrainer} and TinyTrain~\cite{TinyTrain}, also face a similar issue: small batch sizes produce noisier or less reliable gradient estimates, which can compromise the accuracy of their selection criteria. However, since \system{} only relies on the forward pass, we can get around this more easily, as we can run several forward passes, each with a different mini‑batch, and collect all their activations before computing Betti Numbers. For example, with a batch size of 8, we might do 5 passes to gather 40 activation vectors. Since Betti Numbers only use forward activations (no gradients or backpropagation), such "activation pooling" is inexpensive and manageable within the available memory budget. After we choose which layers to train, we switch back to the original small batch size for the actual retraining. 


\subsubsection{Adjustment with Layer Types.} 
In most pre-trained architectures, the base model with pre-trained weights can be broken into additional layers or blocks. These models typically include nested modules that combine both trainable and non-trainable components. \system{} iterates through the model hierarchy, retrieving the activations only from sublayers that contain trainable parameters. Non-trainable layers such as dropout, pooling, or reshaping layers are excluded from consideration. Furthermore, in transformer architectures like ViT, query, key, value, and output linear sublayers within a multi-head self-attention block are treated as a single layer to reduce overall complexity and simplify \system's selection process.

\subsubsection{Balancing Importance and Computation Cost.}\label{sec:normalize} 
Although this is not a universal topological law, it is observed empirically that the first Betti Number, $b_1$ (which measures the number of independent one-dimensional holes in the activation space), tends to be slightly higher in layers with a larger number of activations~\cite{yang2021dive}. Additionally, in convolutional layers, the memory required to store activation values is often significantly greater than that needed to store the layer's parameters. For example, by a factor of about $14\times$ in typical MobileNets~\cite{256kb,tinytl}. To balance the importance selection with the associated computational cost, we normalize the computed first Betti Number, $b^i_1$, for each layer $i$, by the total number of elements in that layer's activations $|a^i|$: $\hat{b}_1^i = \frac{b^i_1}{|a^i|}$.

Although activation size is not a perfect proxy for compute in architectures with heterogeneous kernel sizes, it serves as a consistent and architecture-agnostic normalization factor that correlates with both memory and compute load in most practical DNNs. This normalized first Betti Number $\hat{b}^i_1$ provides a more effective and cost-aware criterion for selection. We note that TinyTrain~\cite{TinyTrain} also normalizes each layer's Fisher score with the number of parameters and number of multiply-accumulate (MAC) operations of that layer, respectively. \figurename~\ref{fig:adabet_detailed} presents a summary of the selection process.

We define a design parameter \(\rho \in [0, 1]\) representing the proportion of layers in \(\mathcal{M}\) selected for training on the target dataset \(\mathbb{D}\). For instance, \(\rho = 0.3\) means that 30\% of the layers, ranked according to their $\hat{b}_1$ values, are chosen for retraining. A value of \(\rho=0 \) corresponds to inference-only use, while \(\rho=1 \) indicates full model retraining. This flexible approach enables \system{} to account for the target device's capabilities, balancing performance gains with computational and memory constraints.

\subsubsection{From Layers to Channels}

\system{} can be naturally extended from layer-level to channel-level selection, enhancing its overall flexibility. Interestingly, our preliminary observations indicate that Betti numbers across channels within the same layer often appear nearly identical. This is likely because channels in deep networks tend to encode similar or correlated feature patterns \cite{channels}, leading to comparable topological structures. Similar to layer selection, we select $\rho_{ch}$ proportion of channels from the selected layers based on their normalized Betti numbers.




Similar to $\rho$ for layer selection, we use \(\rho_{ch} \in [0, 1]\) to represent the proportion of channels to be selected from layers of \(\mathcal{M}\) that were selected in the previous layer selection step. Similar to layer selection, \(\rho_{ch} = 0.3\) means that 30\% of the channels per selected layer, ranked according to their channel-wise $\hat{b}_1$ values, are chosen for retraining. 

\subsubsection{Selective Training on Local Dataset}
\label{sec:training} \system{} performs fine-tuning of the selected subset of layers using the local dataset \(\mathbb{D}\) under resource budgets or on-device. Notice that \system{} can adopt a smaller batch size during such retraining, even if a cumulative larger batch size was used earlier during the layer selection process via activation aggregation (see \S\ref{sec:aggregation}). Moreover, \system{}'s label-free selective training approach is agnostic to the learning paradigm and can be applied in supervised, semi-supervised, or self-supervised settings. While we make no assumptions about the availability of labeled data during retraining, this paper focuses on showcasing the benefits of \system{} in standard supervised retraining scenarios used by our baselines. We leave the investigation of other paradigms for future work.


\section{Evaluation Setup} \label{evaluation}
We compare \system{} with five baselines: Full Training (updating all parameters), Vanilla Transfer Learning (updating only the final layer), Last-K-Layers Learning \cite{last-k} (updating only last K layers), ElasticTrainer~\cite{ElasticTrainer} (dynamic programming-based layer selection), PruneTrain~\cite{PruneTrain} (structured pruning during training), and a Fisher Information-based layer selection approach~\cite{TinyTrain} (selecting the top 10\% layers by Fisher scores). We exclude meta-training and NAS-based methods~\cite{TinyTrain,tinytl,256kb} due to their limited practicality for constrained devices. Experiments are conducted on ViT-B16~\cite{ViT}, ResNet50~\cite{resnet50}, VGG16~\cite{vgg16}, and MobileNetV2~\cite{mobilenetv2}, initialized with ImageNet pre-trained weights, and evaluated on Stanford Dogs~\cite{stanford_dogs}, Oxford-IIIT Pets~\cite{oxford-iiit}, CUB~\cite{CUB}, and Flowers102~\cite{nilsback2008automated}, with images resized to $224\times224$ and standard preprocessing. We use SGDW~\cite{loshchilov2017decoupled} with dataset- and model-specific learning rates, a weight decay of $5\times 10^{-4}$, batch size 8, and set $\rho = 0.1$ and 100\% channel selection for \system{} unless mentioned. Topological computation uses Ripser~\cite{ripser_1,ripser_2}, and memory profiling follows prior benchmarking work~\cite{256kb,tinytl}. Experiments presented in this paper are run on an NVIDIA Tesla V100 (16GB), with additional CPU-based results from Raspberry Pi 4 (with 2 GB RAM) provided in the Appendix \ref{Supp_E}, both exhibiting similar characters. Additional details of the evaluation setup and baselines are presented in the Appendix (\ref{Supp_C} and \ref{Supp_D} respectively). 

\begin{figure*}[t]
\centering
\begin{minipage}[t]{0.75\textwidth}
    \vspace{0pt}
    \centering
    \captionof{table}{Top-1 accuracy of \system{} with $\rho=0.1$ \& $\rho_{ch}=1.0$ and baselines.}
    \centering
\scriptsize

\label{tbl:noise_accuracy}
\resizebox{0.95\textwidth}{!}{%
\begin{tabular}{ccccccccc}
\rowcolor[HTML]{EFEFEF} 
\textbf{}                                                                                 &                                        & \textbf{\begin{tabular}[c]{@{}c@{}}Full \\ Training\end{tabular}} & \textbf{\begin{tabular}[c]{@{}c@{}}Transfer\\ Learning\end{tabular}} &  \textbf{\begin{tabular}[c]{@{}c@{}}Last-K\\ Layers\end{tabular}} & \textbf{\begin{tabular}[c]{@{}c@{}}Fisher \\ Information\end{tabular}} & \textbf{PruneTrain}           & \textbf{\begin{tabular}[c]{@{}c@{}}Elastic \\ Trainer\end{tabular}} & \textbf{\begin{tabular}[c]{@{}c@{}}AdaBet \\ (Ours)\end{tabular}} \\ \hline
                                                                                          & \textbf{ResNet50}                      & 70.87                                                             & 70.01 
                                                                                          & 
                                                                                  72.8        & 73.76                                                                  & 66.09                         & 72.55                                                               & \textbf{75.52}                                                    \\
                                                                                          & \cellcolor[HTML]{EFEFEF}\textbf{VGG16} & \cellcolor[HTML]{EFEFEF}60.82                                     & \cellcolor[HTML]{EFEFEF}61.66                  &      \cellcolor[HTML]{EFEFEF}70.13                             & \cellcolor[HTML]{EFEFEF}70.17                                          & \cellcolor[HTML]{EFEFEF}50.2  & \cellcolor[HTML]{EFEFEF}66.8                                        & \cellcolor[HTML]{EFEFEF}\textbf{72.09}                            \\
                                                                                          & \textbf{MobileNetV2}                   & 80.47                                                             & 80.4  &                              79.62                                  & 80.47                                                                  & 73.31                         & 77.02                                                               & \textbf{81.76}                                                    \\
\multirow{-4}{*}{\begin{tabular}[c]{@{}c@{}}Stanford Dogs\\ (120 classes)\end{tabular}}   & \cellcolor[HTML]{EFEFEF}\textbf{ViT}   & \cellcolor[HTML]{EFEFEF}86.42                                     & \cellcolor[HTML]{EFEFEF}86.16   &            \cellcolor[HTML]{EFEFEF}86.45                           & \cellcolor[HTML]{EFEFEF}14.37                                          & \cellcolor[HTML]{EFEFEF}82.6  & \cellcolor[HTML]{EFEFEF}\textbf{87.79}                              & \cellcolor[HTML]{EFEFEF}85.12                                     \\ \hdashline
                                                                                          & \textbf{ResNet50}                      & 86.18                                                             & 84.63  &                       86.37                                        & 85.94                                                                  & 86.48                         & 87.54                                                               & \textbf{88.74}                                                    \\
                                                                                          & \cellcolor[HTML]{EFEFEF}\textbf{VGG16} & \cellcolor[HTML]{EFEFEF}81.18                                     & \cellcolor[HTML]{EFEFEF}85.37                   &       \cellcolor[HTML]{EFEFEF}85.41              & \cellcolor[HTML]{EFEFEF}\textbf{86.56}                                 & \cellcolor[HTML]{EFEFEF}78.44 & \cellcolor[HTML]{EFEFEF}34.9                                        & \cellcolor[HTML]{EFEFEF}86.14                                     \\
                                                                                          & \textbf{MobileNetV2}                   & 86.95                                                             & 87.78  &                             86.8                                  & 87.16                                                                  & 27.3                          & 86.70                                                               & \textbf{89.93}                                                    \\
\multirow{-4}{*}{\begin{tabular}[c]{@{}c@{}}Oxford IIIT Pets\\ (37 classes)\end{tabular}} & \cellcolor[HTML]{EFEFEF}\textbf{ViT}   & \cellcolor[HTML]{EFEFEF}93.15                                     & \cellcolor[HTML]{EFEFEF}90.21  &                      \cellcolor[HTML]{EFEFEF}88.97                & \cellcolor[HTML]{EFEFEF}26.23                                          & \cellcolor[HTML]{EFEFEF}90.57 & \cellcolor[HTML]{EFEFEF}91.96                                       & \cellcolor[HTML]{EFEFEF}\textbf{92.23}                            \\ \hdashline
                                                                                          & \textbf{ResNet50}                      & 34.76                                                             & 27.13  &                     33.62                                          & 39.85                                                                  & 28.85                         & 24.37                                                               & \textbf{46.37}                                                    \\
                                                                                          & \cellcolor[HTML]{EFEFEF}\textbf{VGG16} & \cellcolor[HTML]{EFEFEF}57.17                                     & \cellcolor[HTML]{EFEFEF}58.37           &             \cellcolor[HTML]{EFEFEF}57.42                & \cellcolor[HTML]{EFEFEF}58.77                                          & \cellcolor[HTML]{EFEFEF}52.77 & \cellcolor[HTML]{EFEFEF}15.75                                       & \cellcolor[HTML]{EFEFEF}\textbf{60.30}                            \\
                                                                                          & \textbf{MobileNetV2}                   & 42.35                                                             & \textbf{50.37}                  &                   38.21                  & 42.26                                                                  & 50.05                         & 15.95                                                               & 46.36                                                             \\
\multirow{-4}{*}{\begin{tabular}[c]{@{}c@{}}CUB\\ (200 classes)\end{tabular}}             & \cellcolor[HTML]{EFEFEF}\textbf{ViT}   & \cellcolor[HTML]{EFEFEF}\textbf{85.73}                            & \cellcolor[HTML]{EFEFEF}80.76               &        \cellcolor[HTML]{EFEFEF}79.89                 & \cellcolor[HTML]{EFEFEF}62.34                                          & \cellcolor[HTML]{EFEFEF}80.22 & \cellcolor[HTML]{EFEFEF}82.93                                       & \cellcolor[HTML]{EFEFEF}79.47                                     \\ \hdashline
                                                                                          & \textbf{ResNet50}                      & 80.56                                                             & 77.18                    &                   79.16                         & 79.05                                                                  & 80.45                         & 80.6                                                                & \textbf{81.39}                                                    \\
                                                                                          & \cellcolor[HTML]{EFEFEF}\textbf{VGG16} & \cellcolor[HTML]{EFEFEF}61.79                                     & \cellcolor[HTML]{EFEFEF}\textbf{75.83}               &      \cellcolor[HTML]{EFEFEF}73.5          & \cellcolor[HTML]{EFEFEF}74.21                                          & \cellcolor[HTML]{EFEFEF}33.68 & \cellcolor[HTML]{EFEFEF}74.86                                       & \cellcolor[HTML]{EFEFEF}74.71                                     \\
                                                                                          & \textbf{MobileNetV2}                   & \textbf{61.72}                                                    & 30.96                   &                    \textbf{65.8}                         & 56.22                                                                  & 37.6                          & 37.6                                                                & 61.5                                                              \\
\multirow{-4}{*}{\begin{tabular}[c]{@{}c@{}}Flowers102\\ (102 classes)\end{tabular}}      & \cellcolor[HTML]{EFEFEF}\textbf{ViT}   & \cellcolor[HTML]{EFEFEF}78.66                                     & \cellcolor[HTML]{EFEFEF}95.45      &          \cellcolor[HTML]{EFEFEF}95.97                        & \cellcolor[HTML]{EFEFEF}13.10                                          & \cellcolor[HTML]{EFEFEF}98.12 & \cellcolor[HTML]{EFEFEF}97.71                                       & \cellcolor[HTML]{EFEFEF}\textbf{98.55}                            \\ \hline
\textbf{Average}                                                                          &                                        & 71.79                                                             & 71.39         &       73.75                                                & 59.41                                                                  & 63.35                         & 64.68                                                               & \textbf{76.26}                                                    \\ \hline
\end{tabular}
} 
\end{minipage}%
\hfill
\begin{minipage}[t]{0.25\textwidth}
    \vspace{-4pt}
    \centering
    \includegraphics[width=0.9\linewidth]{images_mobicom/baseline_sel.pdf}\\
    \vspace{-3pt}
    \includegraphics[width=0.95\linewidth]{images_mobicom/baseline_tpe.pdf}
    \label{fig:time}
    \vspace{-0.4cm}
    \caption{\footnotesize(Top) Time for 1 step of layer selection on Oxford-IIIT Pets. (Bottom) Re-training time for 1 epoch.}
    \label{fig:time}
\end{minipage}
\end{figure*}

\begin{table*}[t]
\centering
\scriptsize
\caption{Peak memory consumption of baselines and \system{} during different stages of the process-layer selection and training. \system{} consumes upto 75.77\% less overall peak memory compared to full training. \system{} also consumes lesser peak memory in comparison to other baselines that require backpropagation for layer selection. Overall peak memory during training for \system{} is comparable to that of inference, transfer learning, and last-k layers.}
\label{tbl:mem_consumption}
\begin{tabular}{lcccclcccc}
\rowcolor[HTML]{EFEFEF} 
\multicolumn{1}{c}{\cellcolor[HTML]{EFEFEF}} & \multicolumn{4}{c}{\cellcolor[HTML]{EFEFEF}\textbf{Peak Selection Memory (MB)}} & \multicolumn{1}{c}{\cellcolor[HTML]{EFEFEF}\textbf{}} & \multicolumn{4}{c}{\cellcolor[HTML]{EFEFEF}\textbf{Peak Training Memory (MB)}} \\
\multicolumn{1}{c}{}                         & \textbf{ResNet50}   & \textbf{ViT}   & \textbf{MobileNetV2}   & \textbf{VGG16}  & \multicolumn{1}{c}{\textbf{}}                         & \textbf{ResNet50}   & \textbf{ViT}   & \textbf{MobileNetV2}  & \textbf{VGG16}  \\ \hline
\textbf{Inference}                           & N/A                   & N/A              & N/A                      & N/A               &                                                       & 243.04              & 86.26          & 315.86                & 932.21          \\
\rowcolor[HTML]{EFEFEF} 
\textbf{Full Training}                       & N/A                   & N/A              & N/A                      & N/A               &                                                       & 777.12              & 386.60         & 373.58                & 988.77          \\ 
\textbf{Transfer Learning}                   & N/A                   & N/A              & N/A                      & N/A               &                                                       & 243.30              & 86.46          & 316.02                & 932.27          \\
\textbf{Last-K Layers}                   & N/A                   & N/A              & N/A                      & N/A               &                                                       & 263.19         & 103.10        & 319.46             &  932.41    \\
\rowcolor[HTML]{EFEFEF}
\textbf{ElasticTrainer}                      & 777.12              & 386.60         & 373.58                 & 988.77          &                                                       & 299.66              & 138.05         & 322.13                & 969.44          \\ 
\textbf{PruneTrain}                          & 777.12              & 386.60         & 373.58                 & 988.77          &                                                       & 777.12              & 386.60         & 373.58                & 988.77          \\
\rowcolor[HTML]{EFEFEF}
\textbf{Fisher Information}                              & 777.12              & 386.60         & 373.58                 & 988.77          &                                                       & 270.98              & 86.27          & 316.02                & 932.28          \\\hline
\textbf{AdaBet (Ours)}                       & 243.04              & 86.26          & 315.86                 & 932.21          &                                                       & 261.78 ($\downarrow$\textbf{66.31\%})      & 93.64 ($\downarrow$\textbf{75.77\%})  & 318.5 ($\downarrow$\textbf{14.74\%})         & 932.45 ($\downarrow$\textbf{5.72\%})   \\
\hline
\end{tabular}
\end{table*}


\section{Experimental Results}\label{sec_results}
In evaluating \system{} against our baselines, we focus on three key aspects: (i) memory and time efficiency (ii) classification performance on the target task, and (iii) the impact of various design choices, including layer and channel proportions, batch size, and normalization. 

\subsection{Classification Accuracy}
\label{sec:accuracy}
\tablename~\ref{tbl:noise_accuracy} presents classification accuracy of \system{} across four target datasets and four DNN architectures, comparing our \system{} with the SOTA baselines. For each configuration, classification accuracy is averaged over three independent runs, totaling 288 runs. Note that all DNNs are pre-trained on the ImageNet dataset, and in this work, the layers or parameters selected by each method are further retrained on the target dataset.

We observe that \system{}  outperforms almost all the baselines across model architectures and datasets. With $\rho = 0.1$, \system{} achieves the highest average accuracy (76.26\%), demonstrating its ability to balance performance while achieving training efficiency (as we report in the next subsection). FI-based approaches suffer significantly as they are often heavily dependent on a separate resource-intensive meta-training phase to boost the accuracy at  runtime~\cite{TinyTrain}. While ElasticTrainer~\cite{ElasticTrainer} performs competitively, \system{} surpasses it across most configurations. Accuracy and loss curves of ResNet50 on Flowers102 for all the baselines and \system{} are given in \ref{Supp_convergence}. Although the CUB dataset shows lower overall performance across methods due to its fine-grained nature, \system{} maintains a competitive accuracy for this dataset as well.

{\bf In summary, \system{} offers a more robust and accurate alternative to existing on-device retraining paradigms, consistently improving performance without the need for full model retraining or compute-expensive gradient-based layer selection. On average, across all 16 dataset–model pairs, \system{} achieves +2.5\% improvement over the second-best baseline.}

\subsection{Resource Efficiency}
\label{sec:efficiency}
\subsubsection{Memory Efficiency}
\tablename~\ref{tbl:mem_consumption} reports peak memory consumption in MB by the selection step and training step separately for all models and baselines. Note that Full Training, Transfer Learning, Last-K Layers \cite{last-k}, and inference do not involve a selection step, and PruneTrain \cite{PruneTrain} selects layers regularly throughout training at frequent intervals. 
Specifically, for ViT, setting $\rho = 0.1$ reduces peak memory usage by nearly 76\% relative to Full Training and ElasticTrainer~\cite{ElasticTrainer}. For ResNet50, the memory reduction is approximately 66\%. Note that these results are independent of the dataset, since all input images are resized to 224×224.

These improvements stem from the fact that \system{} bypasses backpropagation through the entire network, unlike methods like ElasticTrainer and Fisher Information, which must perform at least one complete backward pass, resulting in memory requirements comparable to those of Full Training. Notably, \system{} operates with memory usage comparable to inference and lightweight Transfer Learning yet delivers higher accuracy (as shown in \tablename~\ref{tbl:noise_accuracy}). 

{\bf 
In general, \system{} enables more memory-efficient retraining, achieving nearly a 40\% reduction in peak memory usage on average across four DNNs.}

\subsubsection{Overall Time Efficiency} 
\system{} comprises two stages: selection and then retraining of the selected layers. \system{} achieves significant time efficiency across both stages. As summarized in \tablename~\ref{tbl:related_work}, layer selection approaches differ between the ElasticTrainer and \system{}. To assess the overhead introduced by Betti Number computation compared to the importance calculation and dynamic programming-based selection in ElasticTrainer, we measure the time required for a single selection step in both methods. \figurename~\ref{fig:time}-(top) illustrates the time taken for layer selection on ResNet50 and ViT models using the Oxford-IIIT Pets dataset. Across both architectures, {\bf \system{}'s selection is on average 45\% faster than that of ElasticTrainer\cite{ElasticTrainer} for a single-layer selection step.} In \system{}, time consumption for the selection step is primarily due to the activation accumulation. Notably, since ElasticTrainer requires periodic selection, the time shown will be scaled by the number of times this step is executed during training. Similarly, for Fisher Information, we perform only a single backpropagation, and as shown previously (in \S~\ref{sec:accuracy}), this single backpropagation is often insufficient.

For training, ElasticTrainer~\cite{ElasticTrainer} fixes the number of training epochs to ensure timely convergence. However, this strategy may lead to underfitting depending on the dataset and model. The per epoch training time (in seconds) is reported on Oxford-IIIT Pets with ResNet50 (left) and ViT (right) in \figurename~\ref{fig:time}-(bottom). {\bf \system{} consistently reduces per epoch time compared to Full Training by nearly 11\% (15\% and 7\%, respectively).} Note that Fisher Information shows shorter training times than \system{}, approaching those of Transfer Learning. We observe that, like transfer learning, Fisher Information often selects the last layers (see \figurename~\ref{fig:layer_selection_motivation}(a)-(b)), thereby reducing gradient flow and consequently decreasing training time~\cite{liu2021autofreeze}.

\begin{figure}[t]
    \centering
        \centering
        \includegraphics[width=\linewidth]{images_mobicom/selection_rho.pdf}

    \caption{Layers selected by \system{} at various resource budgets
    }
    \label{fig:selected_layers}
\end{figure}
\begin{figure}[t]
    \centering
    \begin{subfigure}[t]{0.24\textwidth}
        \centering
        \includegraphics[width=\linewidth]{images_mobicom/image_layers.pdf}
        \label{fig:rho}
    \end{subfigure}%
    \begin{subfigure}[t]{0.24\textwidth}
        \centering
        \includegraphics[width=\linewidth]{images_mobicom/image_channels.pdf}
        \label{fig:rho_ch}
    \end{subfigure}%

    \caption{Accuracy and peak memory consumption of ResNet50 in \system{} with (left) varying  $\rho$ on Stanford Dogs and (right) varying percent of channels on Oxford-IIIT Pet datasets. }
    \label{fig:rho_ch}
\end{figure}
\subsection{Ablation Studies on \system}
\label{sec:ablation}
We conduct ablation studies on key \system{} design factors, $\rho$ and $\rho_{ch}$, to evaluate our design choices. Additional experiments on normalization strategies, batch size, and layer selection accumulation size are provided in the Appendix (\ref{Supp_norm}, \ref{Supp_batch}, and \ref{Supp_accumulation}, respectively).



\subsubsection{Selection across Different Budgets} In \figurename~\ref{fig:selected_layers}(a), we show a visualization of the layers selected by \system{} as we vary the resource‐budget parameter $\rho$ across ResNet50 on Oxford-IIIT Pets.
As $\rho$ decreases, \system{} consistently prioritizes a subset of layers, particularly those critical for maintaining performance under constrained budgets. In particular, the same core set of important layers (both early and late) remains selected even under tight budgets. In this way, \system{} identifies the most important layers across models and datasets, yet fine‑tunes its selections to meet predefined resource constraints.

\subsubsection{The Impact of $\rho$} 
We evaluate how varying the layer selection parameter $\rho$, which controls the proportion of layers updated during retraining, affects \system{}’s classification performance and peak memory usage.\figurename~\ref{fig:rho_ch} presents results for ResNet50, plotting classification accuracy on the Stanford Dogs dataset and peak memory consumption as a function of $\rho$. As $\rho$ decreases, \system{} trains a smaller subset of the model layers, leading to a substantial reduction in peak memory usage. For instance, at $\rho = 0.1$, memory consumption drops by approximately 65\% for ResNet50, relative to full-model training ($\rho=1.0$). This confirms that \system{} is an effective approach for reducing the computational overhead for on-device training. In terms of classification accuracy, the models exhibit stability as $\rho$ decreases. Additional results on tuning $\rho$, compared with FI–based and last-$k$ layer selection, are provided in Appendix \ref{E.15}.



    

\subsubsection{The Impact of Channel Selection}
\label{sec:channel}
Extending our analysis beyond layer-level selection, we examine channel-level selection within trainable layers. \figurename~\ref{fig:rho_ch} presents results for ResNet50, highlighting the classification accuracy on the Oxford IIIT Pets dataset and peak memory consumption as a function of $\rho_{ch}$.
Similar to layer selection, decreasing $\rho_{ch}$ trains \system{} on fewer channels, reducing memory usage. For example, at $\rho=0.5$ and $\rho_{ch}=0.1$, ResNet50 memory consumption drops by about 74\% compared to training all channels ($\rho_{ch}=1.0$), while classification accuracy remains stable.

Additionally, in terms of time, as $\rho$ or $\rho_{ch}$ decreases, \system{} selects and trains a smaller subset of the model layers and channels, leading to a reduction in selection and retraining time per epoch relative to full-model training as shown in Appendix \ref{Supp_time}. Overall, \system{} enables a tunable trade-off among memory, time, and accuracy via $\rho$ and $\rho_{ch}$, demonstrating the potential of partial layer/channel selection and retraining for on-device learning.

\section{Conclusion}
We present \system{}, a gradient-free layer- and channel-selection framework for efficiently adapting pre-trained deep models to target local datasets. By leveraging topological features of activation outputs, specifically the first Betti number, \system{} quantifies the representational capacity of network layers and identifies the most informative subsets for retraining. Extensive evaluations across 16 model–dataset pairs demonstrate that \system{} achieves an average 5\% improvement in classification accuracy while reducing peak memory consumption by 40\%, outperforming state-of-the-art baselines. These results establish topological analysis as an effective alternative to gradient-based approaches for efficient adaptation in resource-constrained and privacy-sensitive settings. Future work will focus on extending \system{} toward hardware-aware layer selection by integrating latency and energy profiles; generalizing its applicability to time-series and sensing modalities; analyzing robustness to distribution shifts in real-world deployments; and exploring finer-grained tensor-level topological scoring to enhance flexibility and scalability.

%
%
%
{
    \small
    \bibliographystyle{ieeenat_fullname}
    \bibliography{ref}
}
%


\clearpage
\appendix
\renewcommand\thefigure{\thesection.\arabic{figure}}    
\section{Detailed Related Work}
\label{Supp_A}
The growing need for personalization, data privacy, and post-deployment adaptability underscores the importance of on-device training for DNNs, which necessitates computational resources to perform backpropagation locally, including the calculation and storage of intermediate gradients for updating the DNNs. To minimize resource consumption, existing approaches include techniques such as gradient checkpointing~\cite{gradient_checkpointing,chen2016training}, layer-wise training~\cite{bengio2006greedy}, split learning~\cite {9835178}, and transfer learning~\cite{jiang2022back}. Despite their resource efficiency, these approaches often suffer from reduced accuracy when faced with changes in data or downstream tasks.

Another approach is to co-optimally allocate resources alongside the accuracy of downstream tasks. However, such approaches~\cite{256kb,tinytl,E2Train,Adadrop,10.1145/3437984.3458836} either (1) rely on server-side architecture search to select layers before deployment on the device, or (2) require at least one full backpropagation through the entire model (computing all gradients), which creates a significant bottleneck for resource-constrained devices. 

PruneTrain~\cite{PruneTrain} reduces training parameters through channel-level sparsity and structured pruning. It uses group lasso regularization to shrink the model while maintaining dense computation, enabling dynamic architecture adaptation during training. TinyTrain~\cite{TinyTrain} performs an adaptive parameter selection method based on Fisher Information~(FI) of activations to assess a layer's importance. This allows TinyTrain to be both compute- and memory-efficient on the device. However, to perform the online selection of layers, TinyTrain performs one round of backpropagation through the entire model.
Since relying on a single round of backpropagation may result in suboptimal parameter selection~\cite{FI_ref1,FI_ref2}, TinyTrain performs multiple rounds of backpropagation during a server-side meta-training stage. Meta-training with data from similar distributions might not be realistic, as the distribution of the target private data could change significantly, which negatively impacts the effectiveness of the FI-based layer selection~\cite{FI_ref3}.

ElasticTrainer~\cite{ElasticTrainer} dynamically selects parameters using a dynamic programming approach that minimizes resource consumption of on-device training. Unlike TinyTrain~\cite{TinyTrain}, ElasticTrainer eliminates the need for a meta-training phase by directly selecting important layers on the device. Such a selection process with dynamic programming involves multiple rounds of parameter tuning, each requiring backpropagation through the entire model to ultimately identify the optimal subset of layers. This makes the overall process of selection and training memory-intensive, at times even comparable to that of full training.
A comparison between existing on-device layer selection approaches for DNNs and our proposed system, \system{}, is summarized in~\tablename~\ref{tbl:related_work}.

\section{Background on Betti Numbers}
\label{Supp_B}
In mathematics, {\bf topology} examines the invariant properties of spaces under continuous deformations, focusing on connectedness and compactness. In ML, topology allows us to study the robustness of learned representations to transformations such as scaling, rotation, and noise addition~\cite{topological_features}. In algebraic topology, {\bf homology} analyzes the structure of {\em holes} in a space across various dimensions to understand the shape and connectivity of high-dimensional spaces. Instead of relying on visualizations, homology captures topological features as algebraic invariants, enabling broader and more efficient analysis of complex spaces. 

Formally, the $n^{th}$ homology group of a topological space $\mathcal{X}$, denoted as $H_n(\mathcal{X})$, quantifies {\em $n$-dimensional holes} in $\mathcal{X}$; such that $0$-dimensional holes indicate connected components, 1-dimensional holes indicate loops, and higher-dimensional holes indicate structures such as voids and cavities. 

To calculate $H_n(\mathcal{X})$, we turn the geometric problem (e.g., holes in a space) into an algebraic one (i.e., computing groups) by building a chain complex $C(\mathcal{X})$: a sequence of abelian groups linked by boundary operators $\delta_n$ such that 
$$
\cdots \xrightarrow{\delta_{n+1}} C_{n} \xrightarrow{\delta_{n}} C_{n-1} \xrightarrow{\delta_{n-1}} \cdots 
\xrightarrow{\delta_2} C_1 \xrightarrow{\delta_1} C_0 \xrightarrow{\delta_0} 0.
$$ 
Here, each $\delta_n$ satisfies the key property that consecutive boundaries vanish, i.e., $\delta_n \circ \delta_{n+1} = 0$, meaning that the boundary of a boundary is always zero to make sure that every boundary encloses a well-defined cycle. 
Having $C(\mathcal{X})$, the $n^{th}$ homology group is:
$$
H_n(\mathcal{X}) =Ker(\delta_n)/Im(\delta_{n+1}),
$$
where $\text{Ker}(\delta_n)$ indicates the number of cycles (i.e., elements that map to zero under $\delta_n$), and $\text{Im}(\delta_{n+1})$ indicates the boundaries (i.e., elements that originate from a higher-dimensional space). {\em Thus, $H_n(\mathcal{X})$ captures nontrivial cycles that are not boundaries, effectively identifying the holes in $\mathcal{X}$.} Thanks to the computational technique called {\em persistent homology}, $H_n(\mathcal{X})$ is calculated to indicate how topological features, such as connected components, loops, and voids, emerge and persist across multiple scales in ML and data analysis.

An informative measure that quantifies the number of {\em independent $n$-dimensional holes} in a topological space $\mathcal{X}$ is the  $n^{th}$ {\bf Betti Number} defined as the rank of $H_n(\mathcal{X})$:
\begin{equation}
   b_n = \text{rank}(H_n(\mathcal{X})).
   \label{eqn:betti}
\end{equation} 
For example, if our topological space $\mathcal{X}$ is a circle, we have $b_0 = 1$, because the space is connected, and $b_1 = 1$, because the space contains a single one-dimensional hole (i.e., a loop). In this case, all higher Betti Numbers vanish: $b_n = 0$ for all $n>=2$, as a circle has no higher-dimensional holes. 
For an $n$-dimensional sphere, $b_0 = 1$ indicates a single connected component, and $b_n = 1$, indicates the $n$-dimensional void, while all intermediate Betti Numbers are zero (see Table~\ref{table:betti}). 
\begin{table}[t!]
  \begin{center}
  \captionof{table}{Illustration of Betti Numbers for some common geometrical shapes across different dimensions.}\label{table:betti}
        \label{tbl:bn_basics}
\resizebox{.65\columnwidth}{!}{%
\begin{tabular}{ccccc}
Betti Number &   \includegraphics[width=0.8cm]{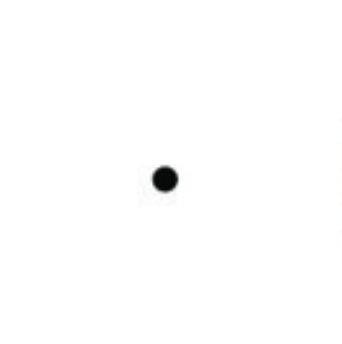}    &     \includegraphics[width=1cm,height=.8cm]{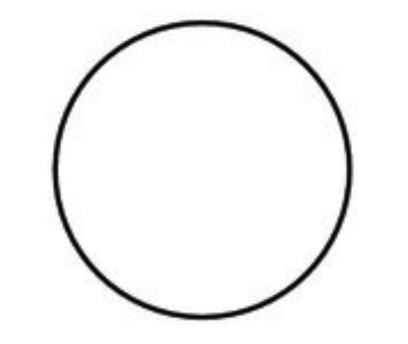}   &  \includegraphics[width=1.cm,height=0.8cm]{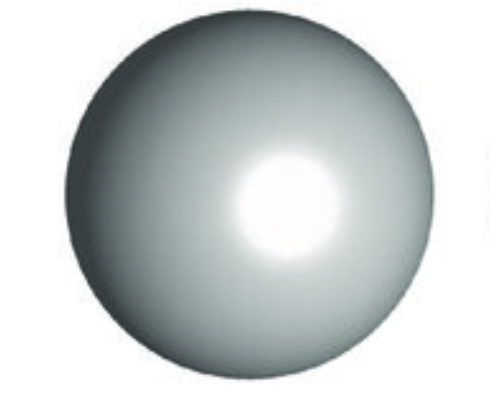}      &  \includegraphics[width=1.cm,height=0.8cm]{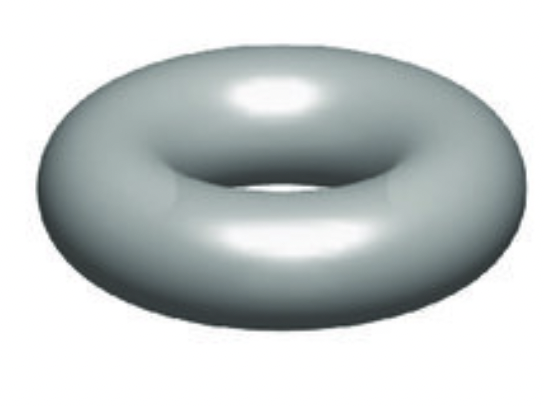}     \\ \hline
\rowcolor[HTML]{EFEFEF} 
$b_0$ & \textbf{1} & \textbf{1} & \textbf{1} & \textbf{1} \\
$b_1$ & 0 & \textbf{1} & 0 & \textbf{2} \\
\rowcolor[HTML]{EFEFEF} 
$b_2$ & 0 & 0 & \textbf{1} & \textbf{1} \\
$b_3$ & 0 & 0 & 0 & 0 \\
\rowcolor[HTML]{EFEFEF} 
:     & : & : & : & : \\
$b_n$ & 0 & 0 & 0 & 0
\end{tabular}
}
  \end{center}
\end{table}
Such a multi-scale perspective, measured through Betti Numbers, helps in identifying topological structures while filtering out noise-induced artifacts. When $\mathcal{X}$ is a torus, persistent homology effectively identifies its two distinct one-dimensional holes and its two-dimensional void; regardless of transformations such as scaling, rotation, or translation. 
\section{Implementation Details}
\label{Supp_C}
{\bf Baselines.}
\label{sec:baselines}
We compare \system{} against five baselines:

\noindent\textit{(1) Full Training}: all layers of the DNN parameters are updated using the local dataset.

\noindent\textit{(2) Vanilla Transfer Learning}: only the final layer is updated, following standard transfer learning approaches~\cite{TL,jiang2022back}.

\noindent\textit{(3) Last-K-Layers}: The last $k$ layers are updated, where $k$ is chosen based on $\rho$ or the percent of layers to be selected.

\noindent\textit{(4) ElasticTrainer}: selected layers are updated during re-training, where selection is done using dynamic programming across multiple rounds of backpropagation~\cite{ElasticTrainer}.

\noindent\textit{(5) PruneTrain}: structured pruning is applied to the DNN during on-device training to reduce training cost~\cite{PruneTrain}.

\noindent\textit{(6) Fisher Information}: layers are selected based on their Fisher Information scores, which are computed from activations and gradients obtained in the first backpropagation round~\cite{TinyTrain}. We select $\rho = 0.1$ (i.e., top 10\%) of layers based on FI scores for re-training; the same default for \system{}. 

Note that baseline (5) differs from TinyTrain~\cite{TinyTrain}, which relies on meta-training. To ensure a fair comparison, we do not compare our approach with server-based meta-training approaches~\cite{TinyTrain,tinytl} or compute-intensive selection methods that require neural architecture search~\cite{256kb}.  We believe these approaches, while valuable, are not practical for constrained devices requiring efficient and timely model retraining. Furthermore, the code for TinyTrain~\cite{TinyTrain} is not available, making it challenging to reproduce their methodology.

{\bf DNN Architectures.}  We evaluate \system{} across four widely-used DNNs: (i) Vision Transformers (ViT-B16) \cite{ViT}, which employs an attention mechanism adopted to vision tasks and consists of 12 Transformer Encoder blocks, (ii) ResNet50 \cite{resnet50}, a deep model comprising 50 convolutional blocks with residual connections, (iii) VGG16 \cite{vgg16}, which includes 13 convolutional blocks with small 3$\times$3 convolution filters, and (iv) MobileNetV2 \cite{mobilenetv2}, a lightweight architecture with inverted residual blocks designed for mobile deployment. All DNNs are initialized with pre-trained weights from the ImageNet dataset. These four architectures were chosen to represent a broad range of design principles. For example, while ViT requires less memory than MobileNetV2, it suffers from slower on-device inference, which makes it more challenging for on-device machine learning~\cite {vit_slow}. 

{\bf Datasets.} As the local dataset \(\mathbb{D}\), we use three datasets: 
(i) Stanford Dogs \cite{stanford_dogs} with 12,000 training and 8,580 testing images of 120 dog breeds, (ii) Oxford-IIIT Pets \cite{oxford-iiit} with 3,680 training and 3,669 testing images from 37 categories of pets, and (iii) CUB  \cite{CUB} with 5,994 training and 5,794 testing images from 200 bird species. (iv) Flowers102~\cite{nilsback2008automated} with 1020 training and 6149 test images from 102 flower categories. To maintain consistency with our primary baseline ElasticTrainer~\cite{ElasticTrainer}, prior to selection and training, we resize all the input images to 224$\times$224 and apply default pre-processing steps such as centering and random flipping.

{\bf Hyperparameters.} For MobileNetV2 model on all datasets we use a learning rate of $1 \times e^{-4}$, for all models on CUB dataset we use a learning rate of $1 \times e^{-2}$, for all other models and datasets we use a learning rate of $1 \times e^{-3}$ and with the SGDW optimizer~\cite{loshchilov2017decoupled}, a cosine decay momentum of 0.9, and a weight decay of $5 \times 10^{-4}$. For all \system{} experiments, we set $\rho = 0.1$, unless otherwise specified. The batch size is set to 8 for all experiments, and the impact of batch size is separately analyzed in Section~\ref{sec:batch_size}. 

\textbf{Libraries and Hardware.} We use TensorFlow 2.15 and TensorFlow Addons 0.23, building on the codebase of ElasticTrainer~\cite{ElasticTrainer}. For topological computations, we employ Ripser 0.6.10~\cite{ripser_1,ripser_2}, which provides fast and memory-efficient calculation of Betti Numbers. Memory computation and profiling are conducted using benchmarking code adapted from~\cite{256kb,tinytl}. Our training and evaluation experiments (\S\ref{sec:efficiency}--\ref{sec:ablation}), unless otherwise specified, are conducted on an NVIDIA Tesla V100 GPU with 16\, GB of VRAM.

\section{More Quantitative Results}
\label{Supp_D}
\begin{figure}[t]
    \centering
    \begin{subfigure}[t]{0.24\textwidth}
        \centering
        \includegraphics[width=\linewidth, keepaspectratio]{images/resnet_acc_norm.pdf}
    \end{subfigure}
    \begin{subfigure}[t]{0.24\textwidth}
        \centering
        \includegraphics[width=\linewidth, keepaspectratio]{images/resnet_mem_norm.pdf}
    \end{subfigure}
    \caption{ (Left) accuracy and (right) peak memory consumption  of \system{} under three normalization strategies applied to Betti Numbers. Activation normalization (our choice) offers the best trade-off between accuracy and memory usage, particularly under tighter selection budgets (\(\rho=0.1\)).}
    \label{fig:norm}
    \label{fig:norm}
\end{figure}
\subsection{The Impact of Normalization.} 
\label{Supp_norm} Betti numbers capture the topological characteristics of activations, reflecting a layer's learning capacity. However, without normalization, layers with a larger number of activations naturally yield higher Betti numbers, making direct comparisons misleading. As discussed before, we normalize the Betti Numbers before layer selection to address this bias (see \S\ref{sec:normalize}). Here, we compare our normalization strategy with two other alternatives: (i) no normalization, and (ii) normalization by the number of trainable parameters (weight size).

\figurename~\ref{fig:norm} compares the accuracy (left) and peak memory consumption (right) of ResNet50 across these normalization strategies for two values of $\rho$ on Stanford Dogs dataset. Normalizing by activation size achieves the best trade-off, leading to higher accuracy and lower memory consumption. In contrast, weight normalization degrades accuracy, particularly at $\rho = 0.1$, while increasing peak memory usage. Activation normalization improves the performance, confirming its suitability for efficient layer selection.

\subsection{Selection across Training Epochs.}
\begin{figure}[t]
        \centering
        \includegraphics[width=\columnwidth]{images_mobicom/training_selection.pdf}
        \caption{Layers selected by \system{} from a model before any training updates (epoch 0), from a partially trained model (epochs 1 and 15), and a trained model (epoch 30)}
        \label{fig:layer_sel_epoch}
\end{figure}
\figurename~\ref{fig:layer_sel_epoch}, illustrates the dynamics of layer selection by \system{} across different training stages on ResNet50 and Oxford-IIIT Pets. At epoch 0 (before any training updates), \system{} already shows a selection pattern similar to epoch 1 and epoch 15. The distribution of selected layers shows that certain layers are consistently selected while others are skipped, indicating that \system{} consistently keeps the focus on layers that contribute more effectively to learning.

\subsection{Impact of Batch Accumulation during Selection.}
\label{Supp_accumulation}
\begin{figure}  
  \centering
  \includegraphics[width=0.75\linewidth, keepaspectratio]{images_mobicom/resnet_batchacc_sel.pdf}
  \caption{Performance of \system{} using VGG16 with increasing number of data batches (each of size 8) accumulated during selection on Oxford-IIIT Pets dataset.}
    \label{fig:selection_batches}
\end{figure}
To counter Betti Number's dependence on the volume of data used for its computation, \system{} uses activation aggregation (see Section \ref{sec:aggregation}). To find the ideal number of data batches required for \system{}, we run an ablation study as shown in \figurename~\ref{fig:selection_batches}. We observe that as the number of batches increases, the performance increases until activations from 5 batches are accumulated for Betti computation and layer selection. Beyond 5, the performance remained almost the same with consistent layer selection. This shows that \system{} requires an aggregated equivalent of 5 batches, each with a batch size of 8, to capture robust layer importance and select layers to achieve optimal performance.
\subsection{Convergence}
\label{Supp_convergence}
Training dynamics are shown in \figurename~\ref{fig:acc_loss_curves}, with test accuracy (left plot) and training loss curves (right plot) over 50 training epochs of ResNet50 on the Oxford-IIIT Pets dataset. ElasticTrainer shows the fastest convergence and ultimately matches the final accuracy achieved by \system{}, while Full Training converges slowly and to a lower accuracy. Transfer Learning, which only updates the final layers, converges slowly and yields the lowest accuracy, highlighting the importance of informed layer selection. Fisher Information performs just better than Transfer Learning.

Although ElasticTrainer initially converges quickly for training loss, \system{}  achieves a lower final loss, indicating a better fit. In contrast, Full Training yields higher loss values, and Transfer Learning shows the slowest reduction with the highest final loss.

{\bf Overall, the training dynamics of \system{} demonstrate a balance between fast convergence, lower loss, and competitive accuracy.}
\begin{figure}[t]
    \centering
    \begin{subfigure}[t]{.5\columnwidth}
        \centering
        \includegraphics[width=\linewidth, keepaspectratio]{images_mobicom/Accuracy_vs_Epochs.pdf}
        \label{fig:test_acc}
    \end{subfigure}%
    \begin{subfigure}[t]{0.5\columnwidth}
        \centering
        \includegraphics[width=\linewidth, keepaspectratio]{images_mobicom/Loss_vs_Epochs.pdf}
        \label{fig:train_loss}
    \end{subfigure}
    \caption{(Left) classification accuracy on test set and (right) training loss of ResNet50 with Flowers102 dataset across baselines and \system{} with $\rho=0.1$. 
    }
    \label{fig:acc_loss_curves}
\end{figure}
\subsection{Impact of Batch Size}
\label{Supp_batch}
\label{sec:batch_size}
\begin{figure}[t]
    \centering
    \begin{subfigure}[t]{0.50\columnwidth}
        \centering
        \includegraphics[width=\linewidth]{images_mobicom/resnet_batch_acc.pdf}
    \end{subfigure}%
    \begin{subfigure}[t]{0.50\columnwidth}
        \centering
        \includegraphics[width=\linewidth]{images_mobicom/resnet_batch_mem.pdf}
    \end{subfigure}
    
    \begin{subfigure}[t]{0.50\columnwidth}
        \centering
        \includegraphics[width=\linewidth]{images_mobicom/resnet_batch_sel.pdf}
    \end{subfigure}%
    \begin{subfigure}[t]{0.50\columnwidth}
        \centering
        \includegraphics[width=\linewidth]{images_mobicom/resnet_batch_tpe.pdf}
    \end{subfigure}
    \caption{Accuracy, peak memory consumption, selection time, and training time per epoch of ResNet50 in \system{} with varying batch size. 
    }
    \label{fig:batches}
\end{figure}
\figurename~\ref{fig:batches} presents a performance study of \system{} under varying batch sizes, evaluating its effect on accuracy, memory usage, selection time, and per-epoch training time. We observe that \system{} maintains high accuracy across all configurations, with only a slight drop as batch size decreases beyond 8. This stability underscores its robustness to batch size variation—an important trait for practical deployment.
Memory consumption, as expected, grows significantly with batch size, with a sharp increase beyond 8. This reinforces the importance of using smaller batch sizes for on-device scenarios, where memory resources are typically constrained.

Interestingly, while Betti Number estimations are generally expected to benefit from larger batch sizes due to better topological representation, we do not observe such a trend here. This is attributed to the accumulation strategy employed in \system{} (see  \S~\ref{sec:aggregation}), wherein activations from multiple smaller batches are aggregated to form an adequate batch size of 40 for layer selection. This allows us to retain the topological richness of larger batches while remaining within device constraints. Notably, the retraining phase continues to use the original smaller batch size.

Selection time across batch sizes decreases steadily until batch size 16, after which the accumulation strategy considers the input batch size as such to form an effective batch of size approximately 40 for selection. Beyond a batch size 16, the accumulation strategy fixes it at 16 to counter the memory consumption with increasing batch size. This effect can be observed at the constant selection time beyond batch size 16. Meanwhile, the time per training epoch decreases steadily with increasing batch size, following standard mini-batch processing behavior. Overall, these results validate that \system{} is compatible with small-batch operation and performs reliably under varying resource budgets, with minimal accuracy degradation and strong efficiency characteristics.

\subsection{Impact of $\rho$ and $\rho_{ch}$ on Time Efficiency.}
\label{Supp_time}
\begin{figure}[h]
    \centering
    \begin{subfigure}[t]{0.23\textwidth}
        \centering
        \includegraphics[width=\linewidth]{images_mobicom/resnet_rho_sel_time.pdf}
    \end{subfigure}
    \begin{subfigure}[t]{0.23\textwidth}
        \centering
        \includegraphics[width=\linewidth]{images_mobicom/resnet_rho_tpe.pdf}
    \end{subfigure}
    \caption{Selection time and Time per epoch of ResNet50 in \system{} with varying $\rho$ on Stanford Dogs datasets. }
    \label{D.13}
\end{figure}

\begin{figure}[h]
    \centering
    \begin{subfigure}[t]{0.23\textwidth}
        \centering
        \includegraphics[width=\linewidth]{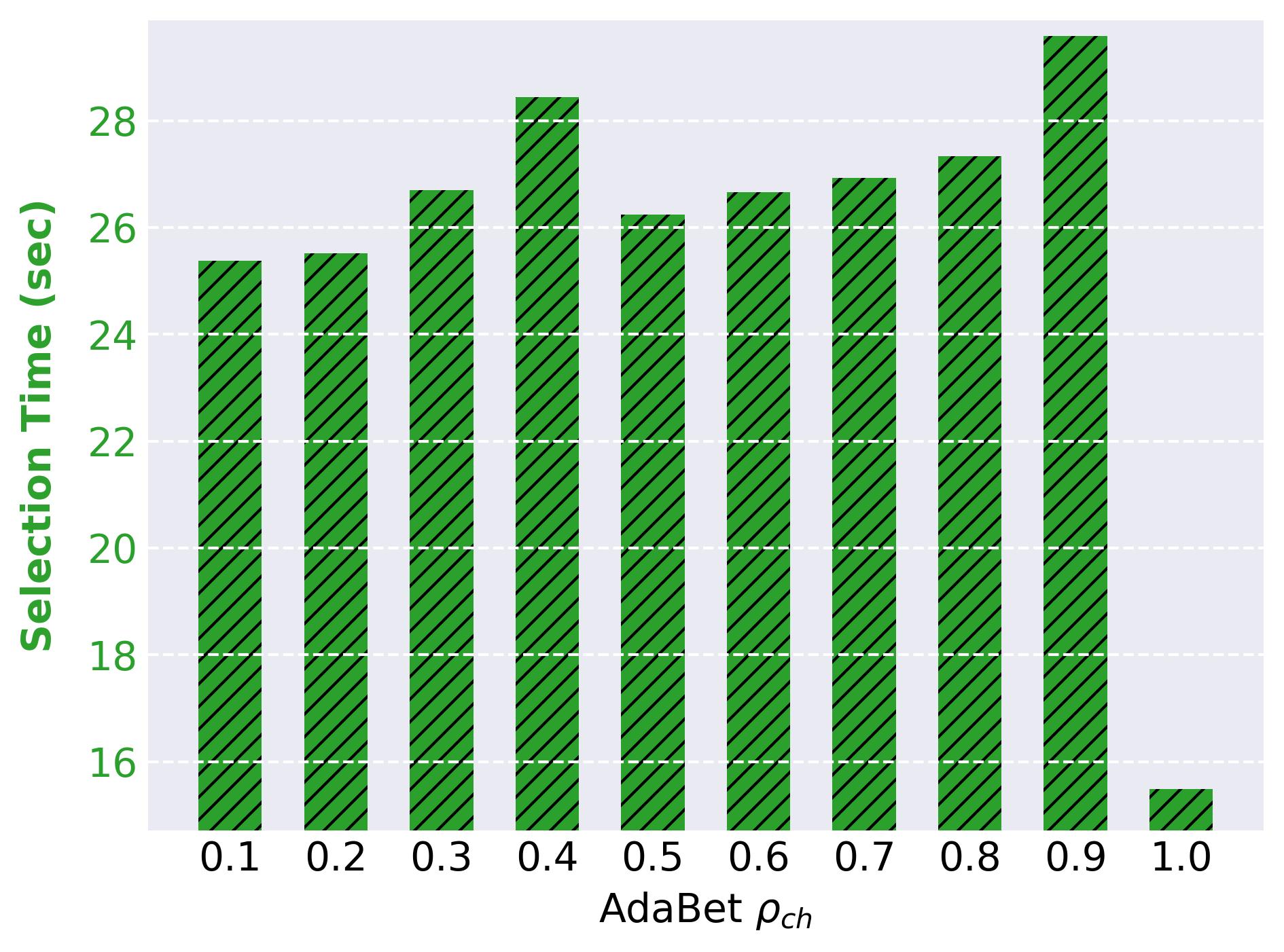}
    \end{subfigure}
    \begin{subfigure}[t]{0.23\textwidth}
        \centering
        \includegraphics[width=\linewidth]{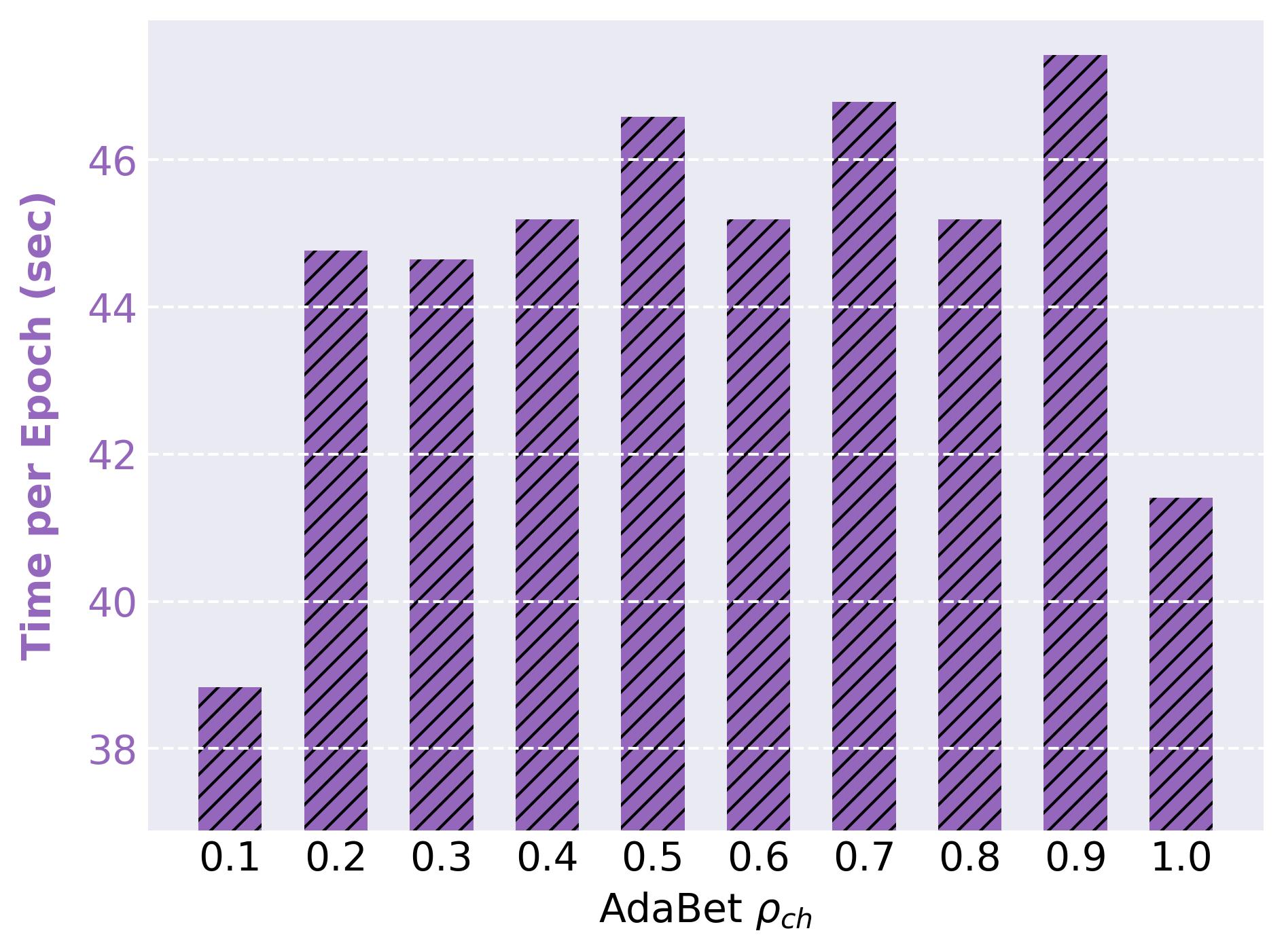}
    \end{subfigure}
    \caption{Selection time and Time per epoch of ResNet50 in \system{} with varying $\rho_{ch}$ on Stanford Dogs datasets. }
    \label{D.14}
\end{figure}

In \system{}, $\rho$ and $\rho_{ch}$ acts as a hyperparameters that choose the proportion layers and channels per layer to be selected for training from the model. This parameter provides a trade-off between performance and peak memory consumption of the model. In this section, we are looking at the time efficiency brought about by \system{} with varying $\rho$ and $\rho_{ch}$ in addition to peak memory and accuracy as shown in \ref{fig:rho} and \ref{fig:rho_ch}. 

In terms of time, as $\rho$ decreases, \system{} selects and trains a smaller subset of the model layers, leading to an approximately 30\% reduction in selection time and approximately 45\% reduction in training time per epoch for ResNet50, relative to full-model training as shown in Fig.\ref{D.13}. However, the decrease in time is less prominent with channel selection. As $\rho_{ch}$ decreases, a smaller subset of layers are selected per layer that was selected in the previous step. A $\rho_{ch}=1$ indicate no layer selection. As shown in Fig.\ref{D.14}, we observe that this consumes least time as expected and with decreasing $\rho_{ch}$, the selection time decreases by about 15\%. Time per epoch decreases by about 6\% as $\rho_{ch}$ decreases from 0.9 to 0.2 with a more significant drop and $\rho_{ch}$ is 0.1. 

\subsection{Controllability of $\rho$}
\begin{figure}[h]
    \centering
    \begin{subfigure}[t]{0.35\textwidth}
        \centering
        \includegraphics[width=\linewidth]{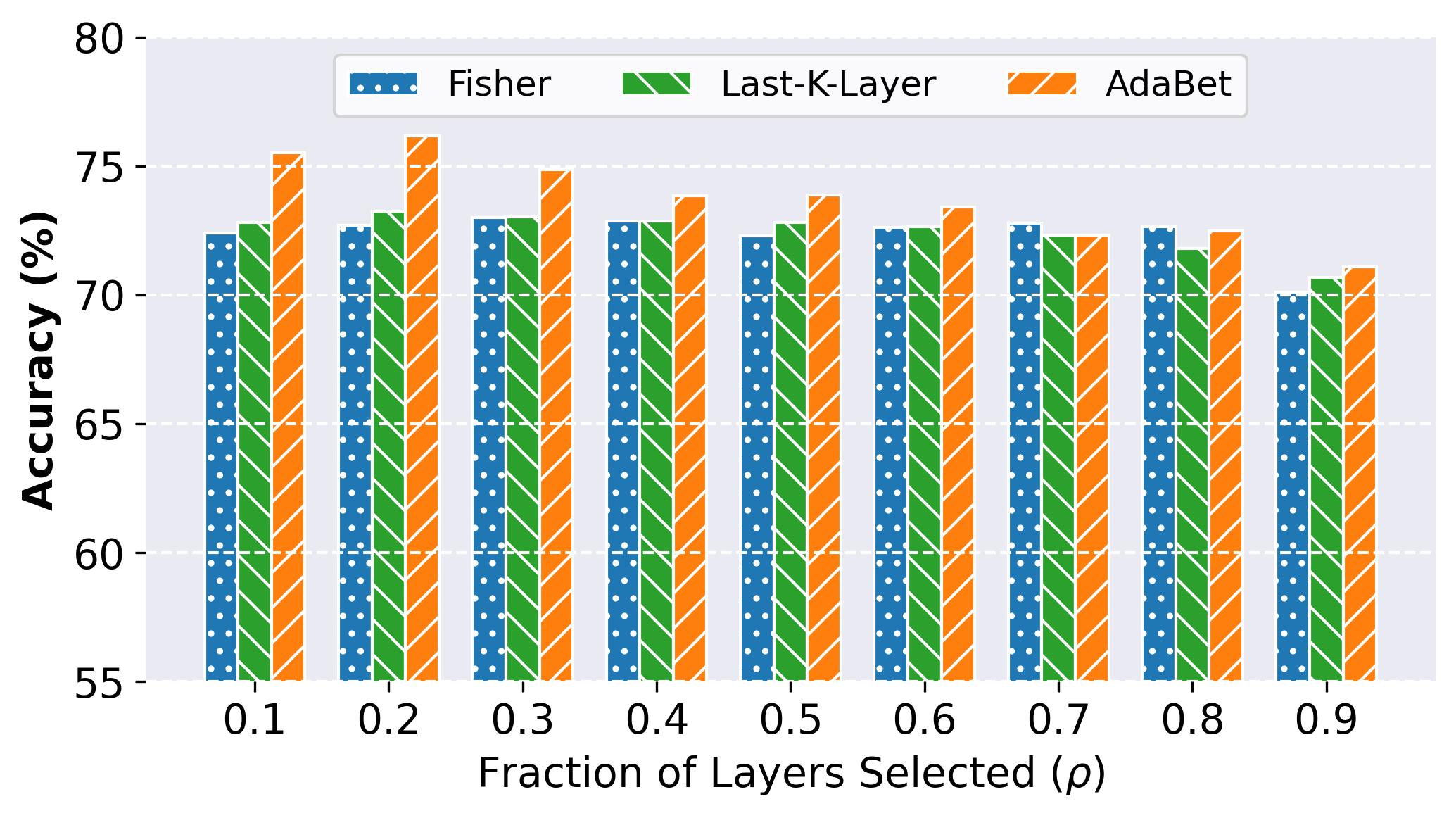}
    \end{subfigure}
    \begin{subfigure}[t]{0.35\textwidth}
        \centering
        \includegraphics[width=\linewidth]{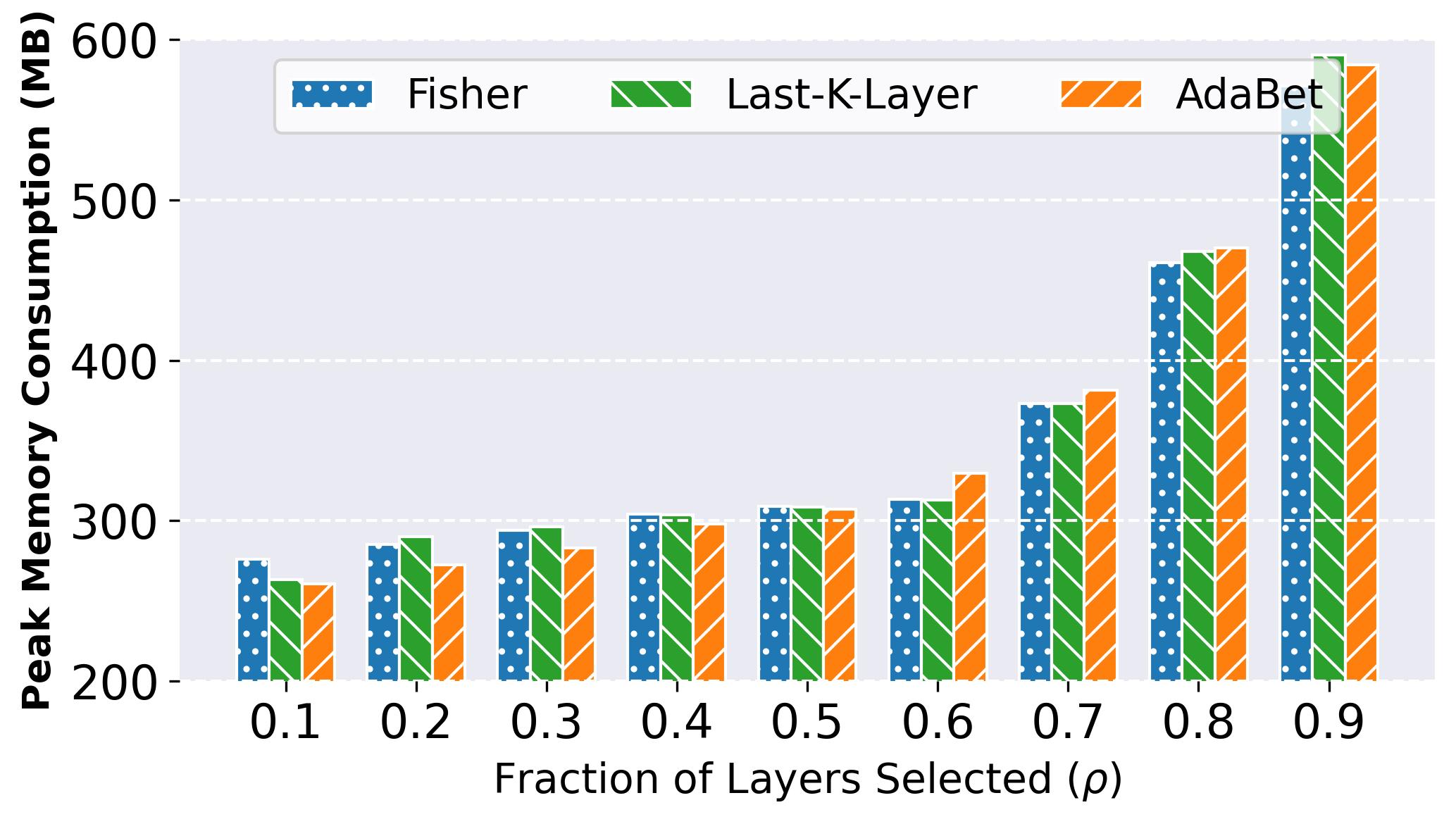}
        
    \end{subfigure}
    \caption{Accuracy(\%) and Peak Memory consumption (MB) of layer selection baselines with ResNet50 on Stanford Dogs.}
    \label{E.15}
\end{figure}

For the same $\rho$, last-$k$-layer training requires comparable or slightly more memory yet consistently underperforms in accuracy, showing that heuristic selections discards learning utility, potentially never updating the key layers responsible for downstream feature extraction. Although accuracy is not strictly monotonic in $\rho$, experiments demonstrate that \system{} never exhibits the failure mode of spending more memory for worse accuracy. Across all $\rho$, it consistently dominates last-k-layer and Fisher Information-based selection at equal or lower peak memory, defining a strictly better accuracy–memory frontier \ref{E.15}. Once $\rho$ is fixed, AdaBet produces a tightly bounded peak-memory footprint that remains well below full fine-tuning while exhibiting better accuracy.

\section{On-Device Results}
\label{Supp_E}
\begin{figure}[h]
    \centering
    \begin{subfigure}[t]{0.23\textwidth}
        \centering
        \includegraphics[width=\linewidth]{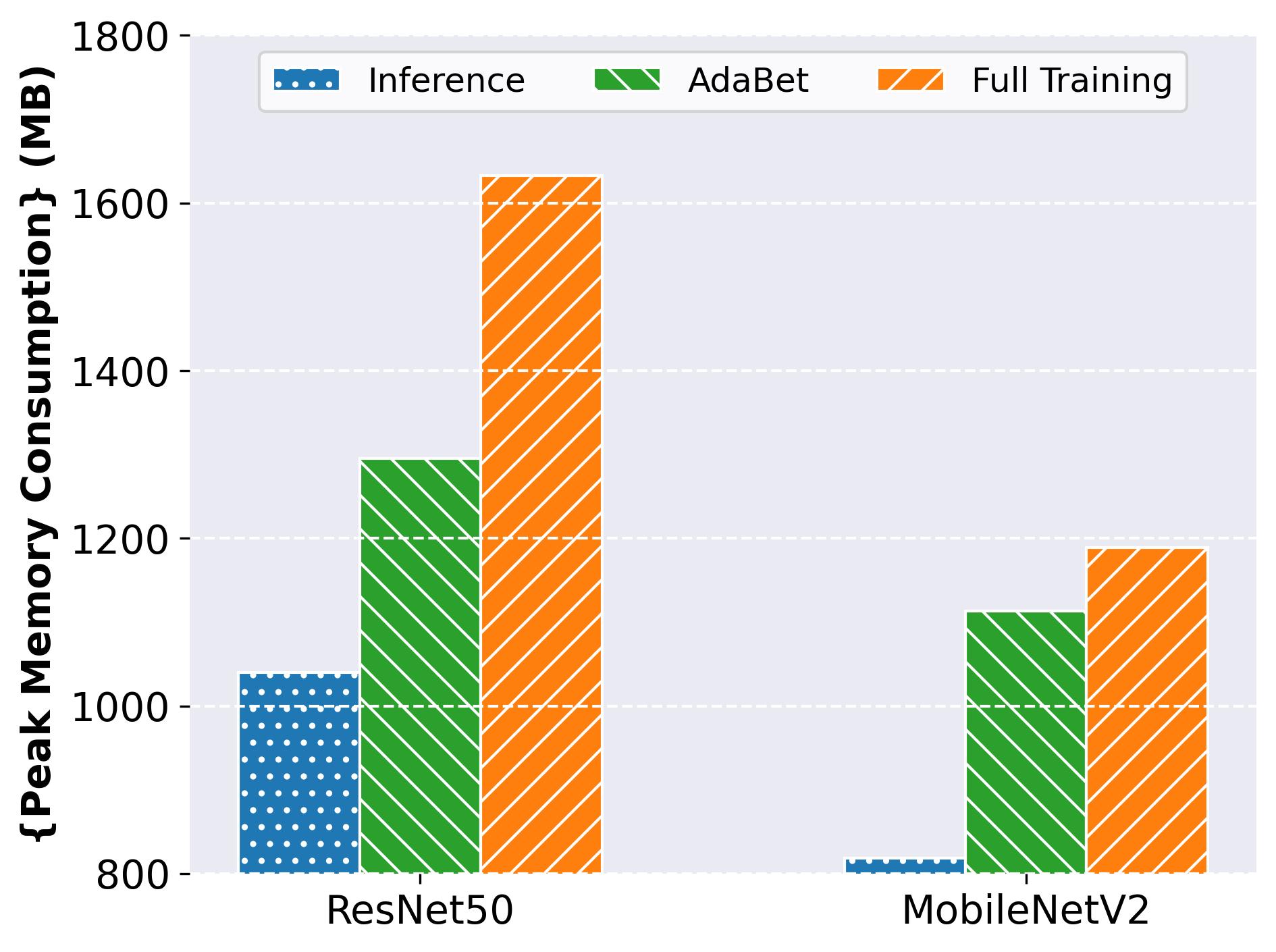}
    \end{subfigure}
    \begin{subfigure}[t]{0.23\textwidth}
        \centering
        \includegraphics[width=\linewidth]{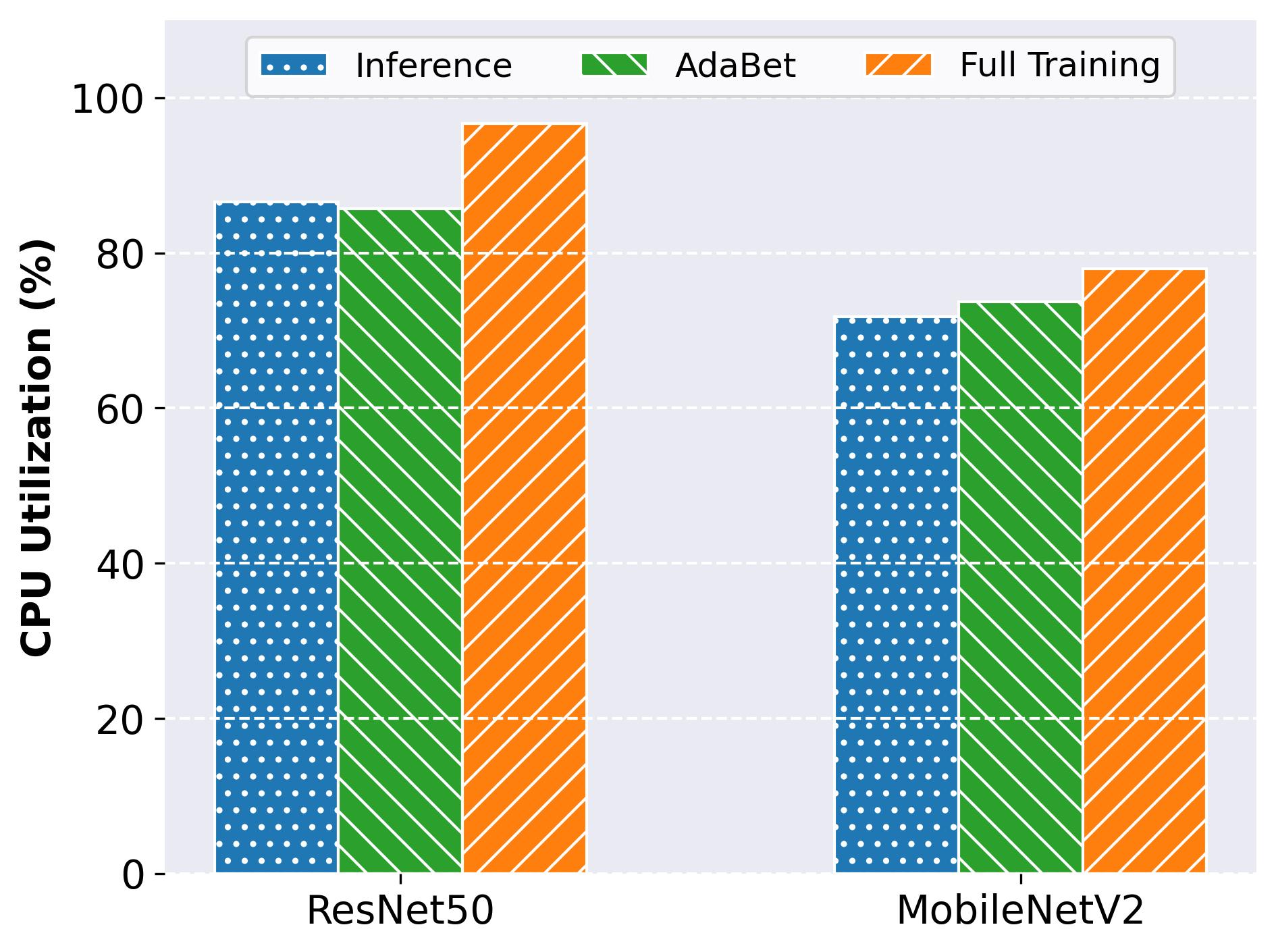}
    \end{subfigure}
    \caption{Peak Memory consumption (MB) and CPU utilization (\%) of ResNet50 and MobileNetV2 on Flowers102 when run on a Raspberry Pi4.}
    \label{E.15}
\end{figure}

For on-device CPU based results we run \system{} on a 2GB Raspberry Pi4. We observe performance, memory consumption, and time consumption similar to that of GPU based results. For the Pi based on-device results, we record peak memory consumption and CPU utilization for full training, \system{}, and inference on ResNet50 and MobilenetV2 - two state-of-the-art architectures for image recognition - on Flowers102 dataset for 102 class image classifiaction task with a batch size of 2, $\rho=0.1$ and $\rho_{ch}=1.0$. All other hyperparameters were maintained the same as GPU as mentioned in \ref{evaluation}. Due to limited compute availability on Raspberry Pi4, we were unable to run any methods on ViT. VGG16 was slow on a Raspberry Pi and we also observed minimal memory or cpu utilization gains. Unlike memory consumption recorded in GPU using methods similar to that in \cite{256kb}, we use total memory consumed by all process run on the CPU (measured using the library "psutils"). Although closer to what is required in real-world deployments, this is often influenced by other CPU processes that may be independent of the training or inference tasks. Similar influence can be expected on CPU utilization. We do not isolate the training process for the purposes of these results. 

Fig. \ref{E.15} shows the peak memory consumption and CPU utilization of the models. As shown, we observe a 20.6\% and 6.2\% reduction in peak memory consumption on \system{} in comparison to full training on ResNet50 and MobileNetV2 respectively. In terms of CPU utilization, we observe a reduction of 11.3\% and 7.9\% respectively for ResNet50 and MobileNetV2  in comparison to full training. \textbf{These results demonstrate that \system{} can be effectively deployed on CPU based devices for efficient DNN tasks. }
\end{document}